\pdfoutput=1

\documentclass[11pt]{article}

\usepackage[final]{acl}

\usepackage{times}
\usepackage{latexsym}
\usepackage{booktabs}
\usepackage{subcaption}
\usepackage{multirow}
\usepackage[T1]{fontenc}
\usepackage{graphicx}
\usepackage{url}
\usepackage{xcolor}

\usepackage[utf8]{inputenc}
\usepackage{algorithm, algorithmic}
\usepackage{amsmath}
\usepackage{microtype}

\usepackage{inconsolata}

\title{Talking Point based Ideological Discourse Analysis in News Events}

\author{
Nishanth Nakshatri\textsuperscript{$\clubsuit$} \quad
Nikhil Mehta\textsuperscript{$\clubsuit$} \quad
    Siyi Liu\textsuperscript{$\diamondsuit$} \quad
    Sihao Chen\textsuperscript{$\diamondsuit$} \\
    \textbf{Daniel J. Hopkins}\textsuperscript{$\diamondsuit$}\quad
    \textbf{Dan Roth}\textsuperscript{$\diamondsuit$}\quad
    \textbf{Dan Goldwasser}\textsuperscript{$\clubsuit$}
    \vspace{0.1in} \\
    \textsuperscript{$\clubsuit$}Purdue University 
    \textsuperscript{$\diamondsuit$}University of Pennsylvania\\
    {\tt \{nnakshat,mehta52,dgoldwas\}@purdue.edu}\\
    {\tt \{siyiliu, sihaoc, danhop, danroth\}@upenn.edu}
}

\begin{document}
\maketitle
\begin{abstract}
Analyzing ideological discourse even in the age of LLMs remains a challenge, as these models often struggle to capture the key elements that shape real-world narratives. Specifically, LLMs fail to focus on characteristic elements driving dominant discourses and lack the ability to integrate contextual information required for understanding abstract ideological views. To address these limitations, we propose a framework motivated by the theory of ideological discourse analysis to analyze news articles related to real-world events. Our framework represents the news articles using a relational structure$-$\textit{talking points}, which captures the interaction between entities, their roles, and media frames along with a topic of discussion. It then constructs a vocabulary of repeating themes$-$\textit{prominent talking points}, that are used to generate ideology-specific viewpoints (or partisan perspectives). We evaluate our framework’s ability to generate these perspectives through automated tasks$-$ideology and partisan classification tasks, supplemented by human validation. Additionally, we demonstrate straightforward applicability of our framework in creating event snapshots, a visual way of interpreting event discourse. We release resulting dataset and model to the community to support further research\footnote{Data available at: \url{https://github.com/nnakshat/TP-IDA}}.

\end{abstract}

%

\section{Introduction}
\begin{figure}[t!]
  \centering
  \includegraphics[width=7.7cm, height=5.2cm]{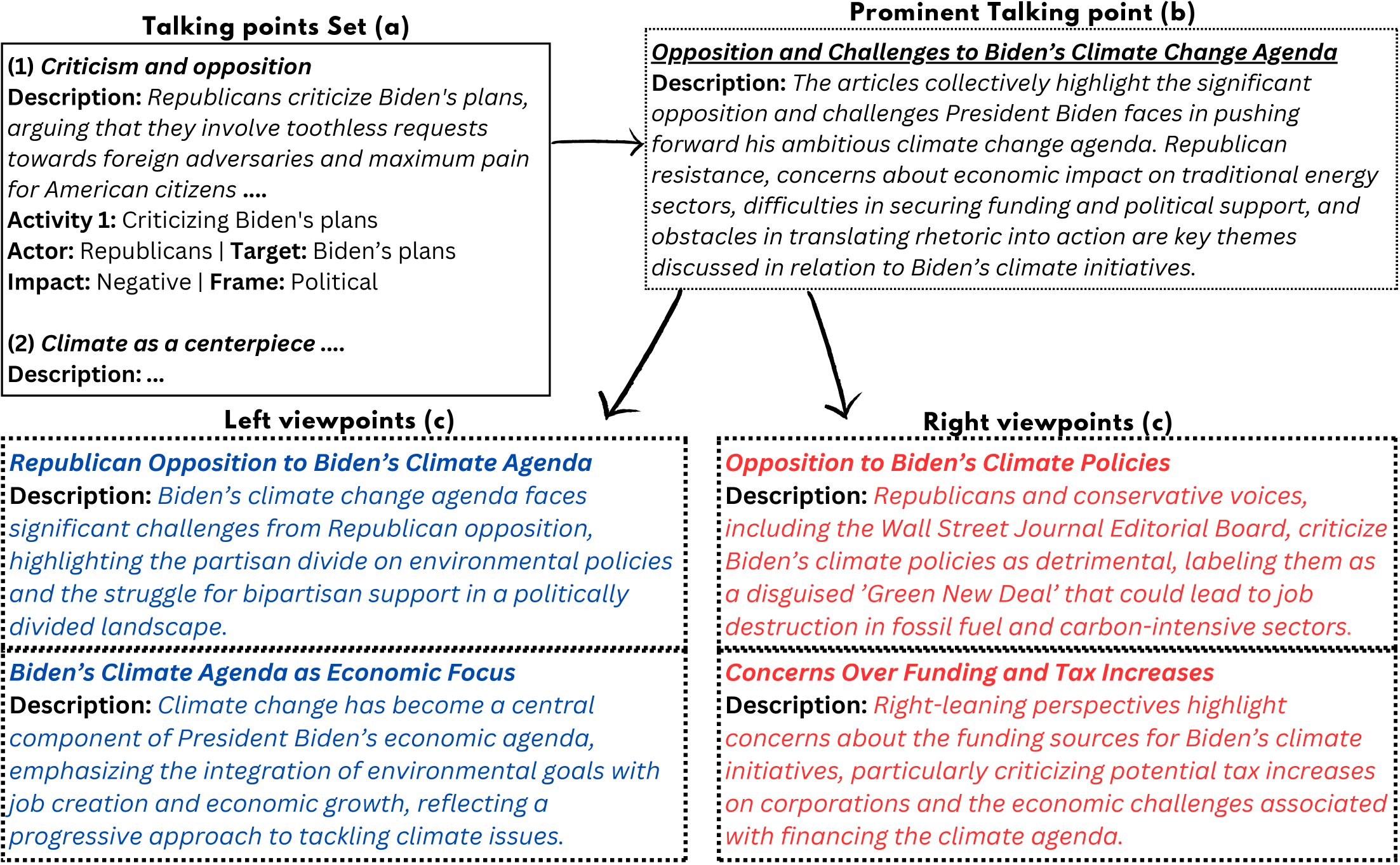}
  \small \caption{\small Shows how each political orientation discusses an event about \textit{Climate Change}. \textbf{(a)} Shows a collection of \textit{talking points} extracted from all the news articles about the event. \textbf{(b)} Shows an example of a \textit{Prominent Talking Point} (\texttt{PTP}), a repeating theme discussed by both political ideologies. \textbf{(c)} Shows \textit{left-} and \textit{right-} ideological interpretations of the \texttt{PTP}.}
  \label{fig:partisan_perspective}
  \vspace{-15pt}
\end{figure}

One of the signs of the growing social and political polarization is the formation of insulated information bubbles~\cite{gentzkow2011ideological,quattrociocchi2016echo,dubois2018echo,garimella2018political}, in which news media discourse is shaped around ideological lines, often intended to shape the readers’ views. Understanding this phenomenon better, to the extent we can examine the degree to which members of the two communities hold opposite accounts of reality, requires computational methods that can compare the narratives of both sides and identify the points in which their accounts converge and diverge (example shown in Figure~\ref{fig:partisan_perspective}).

Past work typically focused on discrete aspects, such as stance and bias detection~\cite{liu2022politics,luo2020detecting,kiesel-etal-2019-semeval,li-goldwasser-2019-encoding}, political news framing~\cite{mendelsohn2021modeling,field2018framing,card-etal-2015-media}, sentiment toward relevant entities~\cite{park2021blames,rashkin2016connotation}, which while relevant, fall short of providing the comprehensive view needed to analyze political discourse. The rise of Large Language Models (LLMs) has a transformative potential for enabling complex discourse analysis connecting these discrete aspects and explaining their relationship. However, realizing it is not straightforward,
%
as demonstrated by several recent works analyzing political texts, either as a straightforward stance prediction task~\cite{ziems2024can}, or mapping political positions to specific stances on policy issues~\cite{santurkar2023whose}. 

\begin{figure*}[h]
\includegraphics[height=4cm, width=14cm]{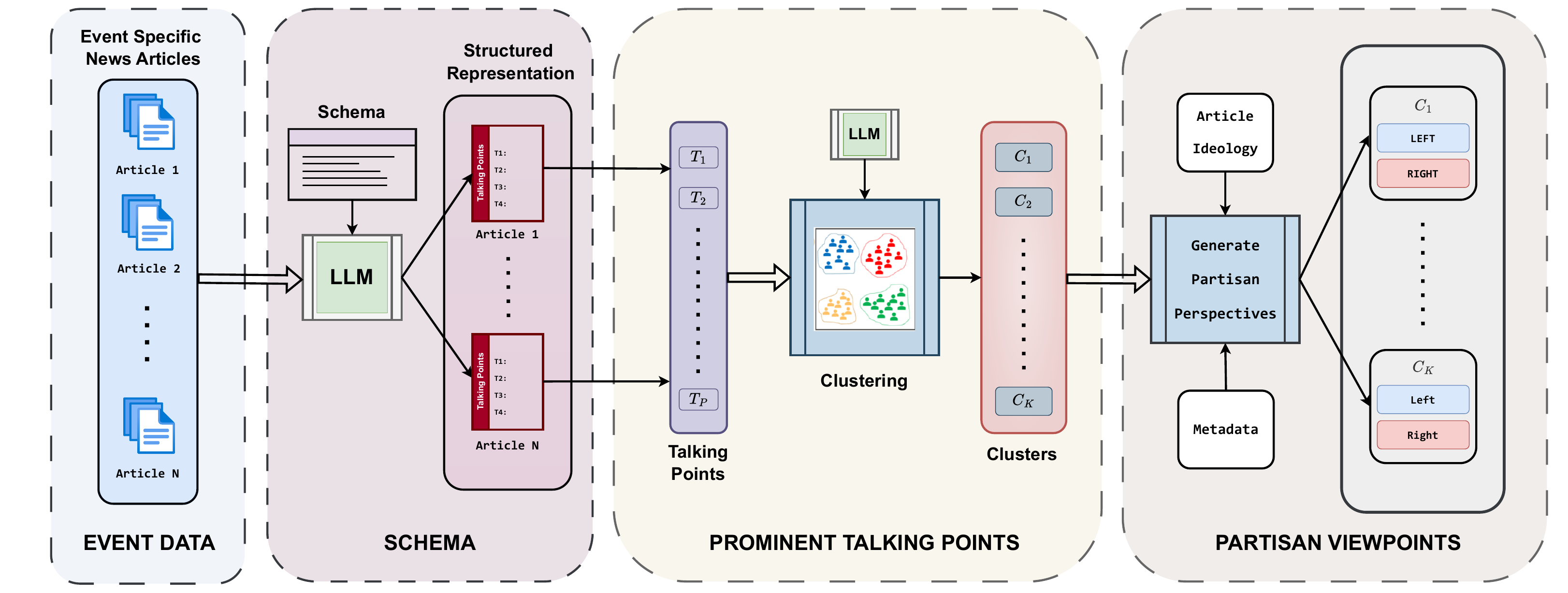}
\centering
\caption{\textbf{Framework overview}: Event-specific news articles are organized into a relational structure, referred to as \textit{talking points}. These are clustered to identify \textit{Prominent Talking Points} (\texttt{PTPs}), a vocabulary of repeating themes that are relevant to the event. Each \texttt{PTP} is infused with ideological information to identify \textit{left-leaning} and \textit{right-leaning} viewpoints, referred to as \textit{partisan perspectives}.}
\label{fig:overallframework}
\vspace{-5pt}
\end{figure*}

To address these challenges, we suggest a structured method to characterize ideological discourse by analyzing repeating themes that may be shared across political sides. We achieve this by defining a relational structure$-$\textit{talking point}, which builds upon foundational research in political discourse analysis~\cite{van1996ideological, van1997political}. 

A \textit{talking point} captures a key aspect of a news article through a short summary and a set of properties. These properties include (1) the lens through which it is discussed, identified using media-frames~\cite{Boydstun2014TrackingTD}, (2) relevant entities mentioned, (3) roles assigned to these entities, and (4) attitudes towards them~\cite{khanehzar-etal-2021-framing,roy2021identifying}. By organizing news articles into this structure, we develop a vocabulary of repeating themes and use these properties to compare variation in discourse across political sides. This approach enables us to detect potential \textit{agenda-setting} attempts~\cite{mccombs1972agenda,scheufele2007framing} by identifying themes overwhelmingly discussed by one side. In addition, it helps uncover \textit{areas of consensus and polarization}, based on examining how both sides engage with the same repeating theme$-$either similarly or in a contrasting manner. Figure~\ref{fig:partisan_narrative_visualization} exemplifies this analysis for Covid-19.

To implement this approach, we develop a pipeline to characterize how each political orientation discusses a \textit{news event} (illustrated in Figure~\ref{fig:overallframework}). First, an LLM extracts main topics of discussion from each event-related article and organizes them according to our structured schema $-$ \textit{talking points}. Next, we identify repeating themes relevant to the event, which we refer to as \textit{Prominent Talking Points} (\texttt{PTPs}). These \texttt{PTPs} are high-level abstractions of frequently repeating \textit{talking points} and are commonly discussed across the political spectrum. To obtain \texttt{PTPs}, we cluster the extracted \textit{talking points} and use an LLM to characterize each cluster. We further refine these clusters by eliminating redundancies and inconsistencies. Once the \texttt{PTPs} are established, we use an LLM to reason about how different political leanings (\textit{left} and \textit{right}) interpret each \texttt{PTP}. The resulting \textit{left-leaning} and \textit{right-leaning} viewpoints are collectively referred to as \textit{partisan perspectives}, capturing differences in framing and attitudes towards entities, as shown in Figure~\ref{fig:partisan_perspective}. Finally, to obtain a comprehensive view of how each political leaning discusses the entire event, we compile the viewpoints associated with each \texttt{PTP}. Specifically, we obtain a \textit{left-leaning perspective} on the event by aggregating all \textit{left-leaning} viewpoints for each \texttt{PTP}, and the \textit{right-leaning perspective} on the event is constructed similarly.

This process generates resources validated by both human and automated techniques, beneficial for researchers. Specifically, we provide a dataset comprising a structured representation of $6,141$ articles on $24$ events related to $4$ contested political issues, sourced from $126$ outlets coded for bias. Further, we also identify \texttt{PTPs} for each event, and provide their respective \textit{partisan perspectives}. 

We suggest two automated methods to evaluate the generated \textit{partisan perspectives}. These evaluations occur at two levels: (1) event-level granularity, and (2) individual \texttt{PTPs}. At the event level, we perform an \textit{ideology classification task} on previously unseen news articles. This task validates whether our method can generate broader partisan discourse that effectively captures the ideological nuances of different political groups. At the \texttt{PTP} level, we perform a \textit{partisan classification task} to measure the degree to which the generated perspectives exhibit political leaning towards a specific ideology.

To summarize, our main contributions are: \textbf{(1)} We propose a new way to characterize ideological discourse in an event via a vocabulary of repeating \textit{talking points}. \textbf{(2)} We suggest an LLM-based pipeline to generate \textit{partisan perspectives}. \textbf{(3)} We perform automated and human validation of generated \textit{partisan perspectives}, resulting in a novel dataset that captures partisan views over multiple events, which can be used to drive future research.


\section{Related Work}
\vspace{-8pt}
Prior work on studying partisan perspectives in NLP has primarily focused on \textit{frames}. While a contested concept, framing is commonly conceived of as a communicative structure in which the speaker highlights specific aspects of an issue to promote a political viewpoint \cite{goffman1974frame, Entman,kinder1998communication,chong2007framing}. \cite{card-etal-2015-media} proposes the Media Frame Corpus that has 15 generic media frames defined by \cite{Boydstun2014TrackingTD}, such as economics or public opinion. In a polarized media environment, frames serve as instrumental mechanisms to promote political agendas through the selective coverage of events (informational bias) and the manipulation of their presentation (lexical bias) \cite{gentzkow2006media, jamieson2007effectiveness, fan-etal-2019-plain}. Prior work has also explored approaches to automatically detect and mitigate framing biases. \cite{liu-etal-2019-detecting, akyurek-etal-2020-multi} identify frames through news headlines, \cite{ji-smith-2017-neural, khanehzar-etal-2021-framing} detect frames at a document level, and \cite{lee2022neus, liu2023opendomain} mitigate framing biases using multi-domain summarization and graphs.
\vspace{-1pt}
However, the formalization of frames oversimplifies the intricacies of partisan perspectives and falls short in capturing the nuance of how political agendas are deliberately conveyed in news articles. In this work, we look closer at news articles, and represent them with a predefined structure of \textit{talking points} $-$ capturing a key aspect of discussion along with interaction between with respective entities, and cluster them to identify repeating themes, to collectively shape the \textit{partisan perspectives}. Identifying \textit{talking points} can be thought of similar to using LLMs to generate explicit representations that helps in assessing arguments~\cite{hoyle2023natural}. Recent work has also explored finer analysis in news articles/political biases. \cite{IJoC6273} presents an entity-focused study of media news framing. \cite{SPINDE2021102505} detects media biases at the word and sentence level, and \cite{Frermann_2023} identifies and uses multi-label frames. Our work complements these by introducing a framework that allows us to establish repeating themes based on talking points to unveil the \textit{partisan perspectives} within an event.




\section{Our Framework}
\label{sec:narrative_framework}
\vspace{-3pt}
In this section, we describe our systematic approach towards generating \textit{partisan perspectives} for a real-world news event. An overview of our approach is illustrated in Figure~\ref{fig:overallframework}.

\subsection{Schema}
\label{subsec:schema}

\paragraph{Background.}Political groups often simplify or obscure their ideological differences. Therefore, to analyze them, theory of ideological discourse analysis, as proposed by~\cite{van1996ideological, van1997political}, introduces semantic discourse structures that can capture the abstract ideological positions of political groups across various social issues and contexts. In particular,~\cite{van1996ideological} suggested that ideological discourse can be predicted by examining local meanings and implications through goal and activity-descriptions, such as questions like “What are our tasks?” and “What are our roles?” etc. Additionally,~\cite{van1997political} proposed to characterize political discourse by defining semantically relevant structures known as \textit{topics}, which are contextualized by their respective topical actors to better understand persuasive communication. 

\paragraph{Structured Representation.} Inspired by these foundational works, rather than directly analyzing news articles, we organize them into a set of \textit{talking points}, a relational structure to characterize ideological differences. 

Similar to the discourse structures in~\cite{van1996ideological, van1997political}, a \textit{talking point} consists of a key discussion element from a news article, which helps to maintain the semantic relevance to the article. To capture nuanced ideological positions and local meanings, we contextualize the discussion by extracting \textit{metadata} that can help analyze ideological differences. This \textit{metadata} is constructed by first identifying the set of entities associated with the discussion. We then capture the relationships between these entities and their influence on each other by identifying the set of activities linked to the discussion. An activity is characterized by the structure \textit{who did what to whom}$-$it includes a brief description, an \textit{actor} who initiates the activity, a \textit{target} affected by the actor, and the sentiment towards the target entity, indicating whether the impact is positive or negative. Additionally, we follow the nomenclature in~\cite{card2015media}, to identify the media frame associated with each activity. Thus, we formally define a \textit{talking point} as a relational structure consisting of a key discussion element from a news article, contextualized with the \textit{metadata} (exemplified in Figure~\ref{fig:partisan_perspective}(a)).
\paragraph{Extraction.} We represent each news article related to a news event using its identified \textit{talking points}. Given a set of $n$ news articles $\{d_z\}_{z=1}^n$ relevant to an event $\mathcal{E}$, our proposed schema distills each article $d_z$ into a set of up to four key \textit{talking points}, denoted as $\{t_i\}_{i=1}^m$, where $m \leq 4$.


Building on~\cite{stammbach-etal-2022-heroes}, which found GPT-3 to be at least $63\%$ accurate in extracting entities and its roles, we prompt an LLM to generate structured representations per our schema. We let the LLM decide the number of \textit{talking points} in a news article. Rather than generating a fixed number of \textit{talking points}, we request the LLM to generate at most $4$ points. This adaptive approach allows the LLM to generate fewer talking points (or no points at all) when there is no useful information in the article. The prompt template is provided in Figure~\ref{fig:promptSchema} in the Appendix.

\subsection{Characterize Partisan Perspectives}
\label{subsec:partisan_perspectives}
After structuring the event-related news articles according to our schema, we outline the methodology used to obtain \textit{left-leaning} and \textit{right-leaning} perspectives on the event $\mathcal{E}$.

\subsubsection{Prominent Talking Points Identification}
\label{subsubsec:clustering}
Although \textit{talking points} are equipped to analyze nuanced political messaging within a news article, they do not directly capture broader ideological discourse at the granularity of a news event $\mathcal{E}$. To capture this, we identify and focus on \textit{Prominent Talking Points} (\texttt{PTPs}) that shape broader narratives. \texttt{PTPs} are repeating themes that are potentially discussed on both sides of the political spectrum.

To obtain \texttt{PTPs}, we first compile a comprehensive talking points set $\mathcal{T} = \{t_s\}_{s=1}^p$ by aggregating all the \textit{talking points} in the articles about $\mathcal{E}$. Then, we cluster $\mathcal{T}$ to identify groups of frequently repeating \textit{talking points}. Each group is labeled using an LLM. The label denotes a higher-level abstraction of the \textit{talking points} within that group. We refer to each label as a \texttt{PTP}, which is a topical representation of the frequently occurring \textit{talking points} that constitute the narrative of an event. Algorithm~\ref{alg:clusteringAlgorithm} outlines the procedure used to obtain \texttt{PTPs}.

\paragraph{Clustering.} We embed each point in $\mathcal{T}$ using a dense retriever model $f$~\cite{ni2021large}, and use HDBSCAN~\cite{campello2013density} to group them into candidate \textit{clusters}. For each cluster $c$, we prompt the LLM to assign a label comprising two components: an \textit{aspect} and a brief \textit{description}. The aspect represents a broad concept evident in the top-$5$ talking points nearest to the cluster centroid, and the description provides a short summary of these points. Note that we exclude metadata during clustering to ensure the clusters accurately capture meaningful topic-based abstractions of \textit{talking points}. Including metadata introduces noise and affects the clarity of these representations.


The initial clustering uses conventional distance metrics and is imperfect. To improve it, we further refine the cluster labels by merging redundant clusters and removing inconsistent ones. Redundancy is addressed by greedily comparing and merging cluster pairs that share the same \textit{aspect}. The resulting refined labels represent the set of \texttt{PTPs} for $\mathcal{E}$. Next, each talking point $t_s \in \mathcal{T}$ is assigned to a \texttt{PTP} based on the cosine similarity between its embedding and the \texttt{PTP's} embedding. This process results in clusters $\{\mathcal{C}_j\}_{j=1}^k$ and their associated labels $\{\mathcal{L}_j\}_{j=1}^k$, referred to as \texttt{PTPs}. More details about the clustering process are in Appendix~\ref{app:membership}.



\begin{algorithm}
\caption{Identify prominent talking points}
\label{alg:clusteringAlgorithm}
\begin{algorithmic}[1]
    \STATE \small \textbf{Input:} Talking points $\mathcal{T} = \{t_s\}_{s=1}^p$
    \STATE \textbf{Initialize: } embeddings $\mathcal{Z} = \{z_s = f(t_s)\}_{s=1}^p\}$, n $\gets$ no. of news articles, $\mathcal{C} \gets \{\}$;
    \WHILE{$|\mathcal{Z}|$ > 0.1$*$ n}
        \STATE clusters $\gets$ \textit{Clustering}($\mathcal{Z}$);
        \STATE labelSet $\gets$ [];
        \FOR{c in clusters}
            \STATE Compute centroid $\mu_c$ by averaging;
            \STATE $Z'$ $\gets$ \textit{getTopKPoints}($c, \mu_c$);
            \STATE cLabel $\gets$ \textit{getClusterLabel($Z'$)};
            \STATE Append cLabel to labelSet;
        \ENDFOR
        \STATE updatedLabels $\gets$ \textit{updateLabelSet}(labelSet);
        \STATE $\mathcal{S}$ $\gets$ \textit{TalkingPtMembership}($\mathcal{T}$, updatedLabels)
        \STATE $T' \gets$ \textit{getClusteredDocs}($\mathcal{S}$)
        \STATE $\mathcal{T}\gets \mathcal{T} \setminus$ $T'$;
        \STATE $\mathcal{Z}\gets \mathcal{Z} \setminus$ \{embeddings of $T'$\};
        \STATE Append clusters in $\mathcal{S}$ to $\mathcal{C}$
        
    \ENDWHILE
    \STATE \textbf{Output:} $k$ clusters $\mathcal{C} = \{\mathcal{C}_j\}_{j=1}^k$ with cluster labels $\{\mathcal{L}_j\}_{j=1}^k$
    
\end{algorithmic}
\end{algorithm}   

\subsubsection{Generate Partisan Perspective}
\label{subsubsection:partisanperspective}
The clustering process identifies a set of \texttt{PTPs} relevant to the event $\mathcal{E}$. To analyze how different political ideologies$-$such as \textit{left} and \textit{right}$-$discuss $\mathcal{E}$, we examine how each ideology engage with these \texttt{PTPs}. Specifically, for each \texttt{PTP}, we define interpretations from the \textit{left} ideology as \textit{left-leaning} viewpoints and interpretations from the \textit{right} ideology as \textit{right-leaning} viewpoints. Collectively, these are referred to as \textit{partisan perspectives}. To construct a comprehensive \textit{left-leaning perspective} on the event $\mathcal{E}$, we aggregate the left-leaning viewpoints associated with each \texttt{PTP}. The \textit{right-leaning perspective} on $\mathcal{E}$ is formed in a similar manner.

\paragraph{Ideological Interpretation of \texttt{PTP}.} \texttt{PTPs} obtained from the clustering process do not inherently capture the ideological specificity needed to define \textit{partisan perspectives}. To address this, we first assign an ideology label $-$ \{\textit{left, right}\}, to each \textit{talking point} within the \texttt{PTP} cluster, denoted as $\mathcal{C}_j$. The ideology label assigned to a \textit{talking point} corresponds to the ideology label of the news article it originates from. Using this labeling, we systematically extract \textit{left-leaning} and \textit{right-leaning} viewpoints for each \texttt{PTP} based on the metadata associated with the \textit{talking points}.

Particularly, to construct a \textit{partisan perspective}, we illustrate the generation of \textit{left-leaning} viewpoints (\textit{right-leaning} ones follow an analogous process). For each \texttt{PTP}, our goal is to generate viewpoints that clearly distinguish the \textit{left-}view from the \textit{right}. To achieve this, we instruct the LLM to create \textit{left-leaning} viewpoints in a contrastive manner. Specifically, we prompt the LLM with characteristic \textit{talking points}: the top-$K$ left-biased points alongside the top-$M$ right-biased ones.\footnote{Based on empirical validation, we set $M=3, K=5$.} This allows the model to highlight contrasts and identify defining features of the \textit{left-}viewpoint. The characteristic \textit{talking points} are selected based on their cosine similarity to the \texttt{PTP}. Note that these \textit{talking points} include metadata such as key actors, their targets, sentiments toward these targets, and the media frame. Incorporating these details ensures that generated \textit{left-leaning} viewpoints capture nuanced relationships between the involved entities.

However, simply using the top-$K$ left-biased \textit{talking points} is not enough, as these may not capture the broader ideological bias in the entire article. To mitigate this, we also include article summaries associated with top-$K$ points in the prompt. To ensure that these summaries reflect ideological biases and relevant information from the \textit{talking points}, we instruct the LLM to generate summaries conditioned on: (1) ideology label of the article (2) \textit{title} associated with the \texttt{PTP}. A detailed prompt template for generating these viewpoints is shown in Figure~\ref{fig:promptSchemaPartisanSumm} in the Appendix.

\paragraph{Metadata Aggregation for each \texttt{PTP}.} Beyond \textit{partisan perspectives}, for each ideology, metadata from the top 50\% of \textit{talking points} is aggregated to identify frequent and distinctive entity pairs with their sentiments. Specifically, we extract the top-3 targets with positive and negative sentiments, their most common associated actors, and the dominant media frame for each actor-target pair. This aggregated view supplements \textit{partisan perspectives} by offering insight into broader dynamics of clusters. 

\begin{figure}[t!]
  \centering
  \includegraphics[width=6.5cm, height=5.6cm]{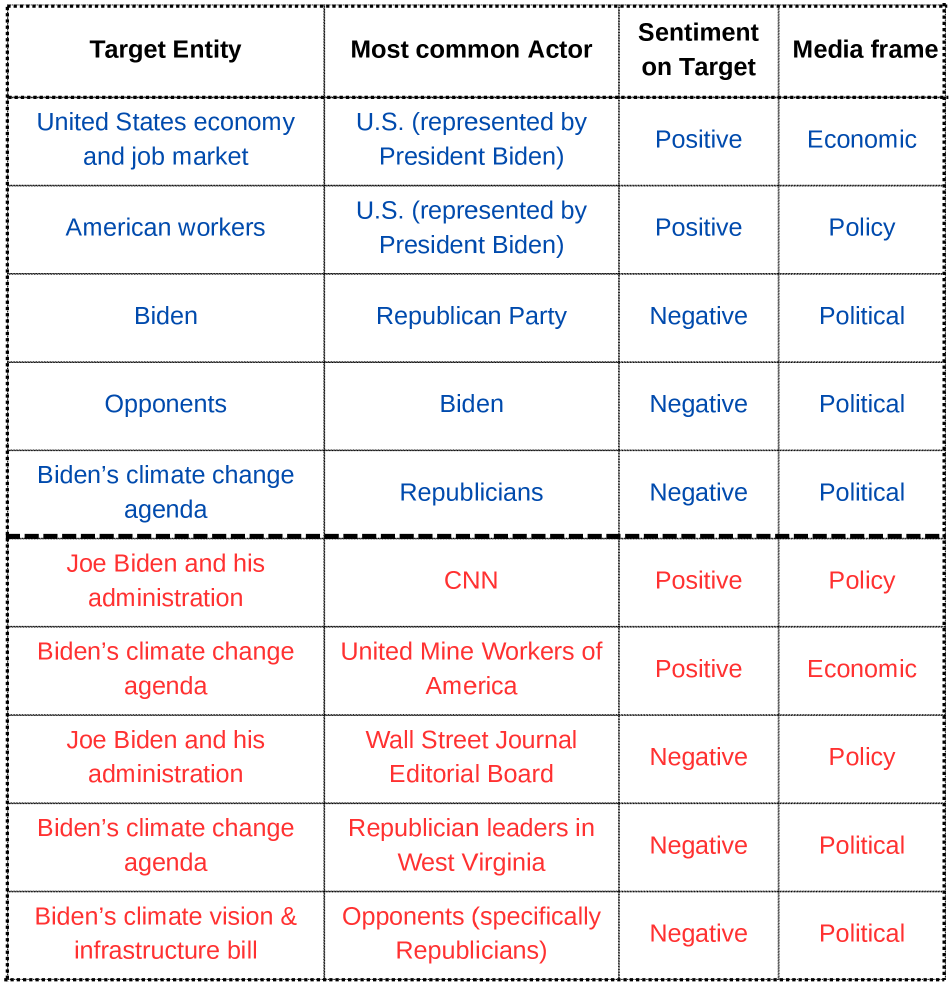}
  \small \caption{\small Shows Aggregated metadata corresponding to the \texttt{PTP} shown in Figure~\ref{fig:partisan_perspective}. Each target is paired with its most common actor, sentiment, and media frame within the actor-target context. Metadata is color-coded: \textcolor{blue}{Blue} for \textit{left} ideology and \textcolor{red}{Red} for \textit{right} ideology.}
  \label{fig:metadata_aggregation_example}
  \vspace{-5pt}
\end{figure}

An example of left- and right-viewpoints for a \texttt{PTP} is shown in Figure~\ref{fig:partisan_perspective}, with its aggregated metadata provided in Figure~\ref{fig:metadata_aggregation_example}. Interestingly, few right-biased sources like \textit{Washington Times} highlight support for \textit{Biden's Climate Agenda} from \textit{United Mine Workers}. This support is based on the agenda's focus on securing jobs for displaced miners.

\vspace{-2pt}
\section{Dataset}
\vspace{-2pt}
To illustrate the effectiveness of our proposed framework, we use the \texttt{KeyEvents} dataset~\cite{nakshatri2023using}. This dataset is constructed by segmenting the archive of news articles from NELA-2021~\cite{nela:DVN/RBKVBM_2022} into a set of temporally motivated news events. To construct these events, \citet{nakshatri2023using} dynamically analyzed the temporal trend of news articles published for a given issue, and identified the temporal landmarks that could signify the presence of an important news event. Then, the news articles published in and around the temporal landmarks were clustered to identify all the documents relevant to the news event. In this manner,~\citet{nakshatri2023using} proposed a dataset comprising of $40k$ news articles with $611$ key news events from $11$ issues.

As our goal is to analyze political discourse and characterize partisan perspectives at an event-level granularity, this dataset can be directly applicable to test the efficacy of our framework. Thus, we manually sample a set of six issues and a set of events which have the highest number of news articles per event from this dataset. Table~\ref{tab:dataset} shows the detailed statistics of our final dataset.


\begin{table}[]
\centering
\small
\begin{tabular}{c|c|c}
\hline
\multicolumn{1}{c|}{\textbf{Issue}} & \textbf{No. of Articles} & \textbf{No. of Events} \\ \hline
\multicolumn{1}{l|}{Climate Change}       & 579            & 8           \\
\multicolumn{1}{l|}{Capitol Insurrection} & 1,609          & 4           \\
\multicolumn{1}{l|}{Immigration}          & 1,137          & 4           \\
\multicolumn{1}{l|}{Coronavirus}          & 2,816          & 8           \\ \hline
\multicolumn{1}{l|}{\textbf{Total Count}}& 6,141&24\\ \hline
\end{tabular}
\small \caption{\small The dataset we use for testing our proposed framework. It is sampled from \citet{nakshatri2023using}.}
\label{tab:dataset}
\vspace{-10pt}
\end{table}

\section{Evaluation}
\vspace{-3pt}
Our goal is to evaluate how effectively our framework generates \textit{partisan perspectives}. To measure its performance, we use two types of automated evaluations (described below) along with human assessments. Additional evaluation of intermediate framework steps is detailed in the Appendix~\ref{app:ptp-eval-task}.

\textbf{Ideology Classification.} In this evaluation, we assess how effectively our framework characterizes ideological discourse at the event level. Specifically, we examine whether the generated perspectives capture event-level political discourse and encode nuanced ideological cues. To achieve this, we consider an \textit{ideology classification task} over previously \textit{unseen} news articles related to the event $\mathcal{E}$. Here, we use \textbf{\textit{only}} the generated \textit{left-leaning} and \textit{right-leaning} perspectives of $\mathcal{E}$ to predict the ideological stance of the article. A strong performance on this task would indicate that the generated perspectives are event-centric rather than being overly specific to individual news articles. Furthermore, it would demonstrate that these perspectives effectively capture the ideological characteristics of different political sides. Our objective is not to outperform existing classification baselines, but rather to determine whether the generated perspectives provide a faithful representation of the broader ideological discourse surrounding an event. 

\textbf{Partisan Classification.} While ideology classification evaluates perspectives at the event level, this task focuses on assessing their quality at the \texttt{PTP} cluster level (as defined in Section~\ref{subsubsection:partisanperspective}). Within each \texttt{PTP} cluster, \textit{left-leaning} viewpoints should align with how \textit{left-} political ideology interprets the issue relative to the \texttt{PTP}, and the same applies to \textit{right-leaning} viewpoints. As a consequence, \textit{left-leaning} viewpoints should be primarily supported by \textit{left-leaning} news articles and not appear in \textit{right-leaning} articles, and vice-versa. This evaluation measures how well the generated perspectives follow this expected pattern.
\vspace{-2pt}
\section{Experiments \& Results}
\vspace{-3pt}
In this section, we discuss the experimental details and findings linked to both automated and human evaluations. We use ChatGPT~\footnote{gpt-3.5-turbo-0125 \cite{openai2022}} as the LLM to obtain \textit{partisan perspectives}, following the approach detailed in Section~\ref{sec:narrative_framework}. This process yields structured representations for articles from every event in our dataset, and we release these, along with the original dataset, to the community.

\subsection{Ideology Classification}
\label{subsec:ideology-classification-task-full}
\textbf{Task Setup.} Given an \textit{unseen} news article related to the event but not part of the initial dataset, the task is to predict its ideology label ({\textit{left, right}}). The central idea involves using \textbf{\textit{only}} the generated \textit{partisan perspectives} for classification. To do this, first, we consider the comprehensive \textit{left-leaning} and \textit{right-leaning perspective} on $\mathcal{E}$. Then, we identify the three most similar \textit{left-} and \textit{right-leaning} viewpoints to the input article using cosine similarity. We denote this as \textit{TopK Event Partisan View}. For classification, we provide the input article along with the \textit{TopK Event Partisan View} to an LLM and prompt it to assign the article to the viewpoint it aligns with most closely. In this process, we replace the explicit labels \textit{\{left-, right-\}} viewpoints with neutral placeholders$-$\textit{\{summary1, summary2\}}, to avoid any inherent bias of LLM. This ensures that the classification relies only on the LLM's ability to align the article with the closest summary. For this task, we curated $481$ unseen event-related articles (see Appendix~\ref{app:datasetExtraction} for details). 


\paragraph{Comparison Methods.} We benchmark task performance with two LLMs$-$ChatGPT and LLAMA3\footnote{meta-llama/Meta-Llama-3-8B-Instruct}~\cite{dubey2024llama3herdmodels}, against these methods: \textbf{1. Direct Prompting}, where the LLM is directly prompted for the article's ideology label. \textbf{2. TopK Event Partisan View}, and \textbf{3. TopK Event Partisan View + Metadata}, which adds metadata to the topK view. For each ideology, metadata is aggregated from the top 50\% of \textit{talking points} closest to \texttt{PTP} (obtained from Section~\ref{subsubsection:partisanperspective}). 

\paragraph{Results.} Table~\ref{tab:ideology-classification-stats} presents ideology classification task results. \textit{TopK Event Partisan View} outperforms the direct prompting baseline, improving F1-scores by $+4.5$ for ChatGPT and $+1$ for LLama. This suggests that the partisan perspectives effectively capture event-level partisan signals that are essential for predicting the ideology of an unseen article. In addition, we observe that the inclusion of metadata can help improve classification task performance. This improvement suggests that metadata produced by our method aids in ideological discourse understanding by capturing the relationship between entities involved.  

\subsubsection{Additional validation of perspectives}
\paragraph{Narrative-LLAMA.} We further validate the effectiveness of generated perspectives by investigating if we could directly fine-tune an LLM to identify event-level discourse signals from individual news articles. Given a news article and its closest ideological viewpoints, the task is to train the model to generate these viewpoints without relying on intermediate steps in our framework. To achieve this, we used the \textit{partisan perspectives} generated by our framework as \textit{training data} to fine-tune a LLAMA3-8B-instruct model using PEFT~\footnote{https://github.com/huggingface/peft} and LORA~\cite{hu2021lora}. We compiled $1100$ examples pairing articles with their \textit{left-} or \textit{right-}viewpoints from \texttt{PTP} cluster, and trained the model using DPO~\cite{rafailov2024directpreferenceoptimizationlanguage} (model training details can be found in Appendix~\ref{app:narrative-llama-training-details}). To evaluate the fine-tuned model, we tested it on ideology classification task using \textit{unseen} articles. We assess classification via: \textbf{Prompting with Generated Perspective}, obtaining perspectives with Narrative-LLAMA and prompting for article ideology.

\paragraph{Results.} From Table~\ref{tab:narrative-llama-performance}, it is evident that the fine-tuned Narrative-LLAMA outperforms the LLAMA3-8b-instruct model by increasing the F1 score by $+3.6$ points in the ideology classification task. This suggests that (1) \textit{partisan perspectives} from our framework that was used as \textit{training data} encodes ideology-specific nuances. (2) generated perspectives capture broader event-level signals than focusing on specific details of the article (as illustrated in Figure~\ref{fig:narrative-llama-qualitative-eval} in the Appendix). Furthermore, using the Llama3-70B-instruct model, we assess the quality of generated perspectives by verifying if they actually appear in the original news articles (Appendix~\ref{app:ideology-classification-task}). Notably, hallucination rates were found to be below $6\%$, and we intend to release the model parameters to the community.

\begin{table}[]
\small
\centering
\resizebox{\columnwidth}{!}{%
\begin{tabular}{|l|c|c|c|l|}
\hline
\multicolumn{1}{|c|}{\textbf{LLM}} & \textbf{Method} & \textbf{Prec.} & \textbf{Recall} & \textbf{F1} \\ \hline
\textbf{LLAMA} & \begin{tabular}[c]{@{}c@{}}Prompting w \\ Generated Perspective\end{tabular} & 87.8 & 84.9 & 84.9 \\ \hline
\textbf{Narrative-LLAMA} & \begin{tabular}[c]{@{}c@{}} Prompting w \\ Generated Perspective\end{tabular} & 89.5 & 88.4 & \textbf{88.5} \\ \hline
\end{tabular}%
}
\small \caption{\small Narrative-LLAMA (fine-tuned model) improves over Llama3 ($+3.6$ F1-score) in ideology classification.}
\label{tab:narrative-llama-performance}
\end{table}

\begin{table}[]
\centering
\resizebox{\columnwidth}{!}{%
\begin{tabular}{c|l|ccl}
\hline
\multirow{2}{*}{\textbf{LLM}} & \multicolumn{1}{c|}{\multirow{2}{*}{\textbf{Method}}} & \multirow{2}{*}{\textbf{Prec.}} & \multirow{2}{*}{\textbf{Recall}} & \multicolumn{1}{c}{\multirow{2}{*}{\textbf{F1}}} \\
 & \multicolumn{1}{c|}{} &  &  & \multicolumn{1}{c}{} \\ \hline
\multirow{3}{*}{\textbf{\begin{tabular}[c]{@{}c@{}}ChatGPT\\ (0-shot)\end{tabular}}} & Direct Prompting & 81.38 & 74.66 & 73.52 \\
 & TopK Event Partisan View & 77.12 & 76.64 & 76.61 \\
 & +Metadata & 81.20 & 79.78 & \textbf{79.69} \\ \hline
\multirow{3}{*}{\textbf{\begin{tabular}[c]{@{}c@{}}ChatGPT\\ (2-shot)\end{tabular}}} & \begin{tabular}[c]{@{}l@{}}Direct Prompting\end{tabular} & 80.47 & 78.23 & 78.09 \\
 & TopK Event Partisan View & 83.02 & 82.61 & \textbf{82.65} \\
 & +Metadata & 83.79 & 82.50 & 82.52 \\ \hline
\multirow{3}{*}{\textbf{\begin{tabular}[c]{@{}c@{}}LLAMA\\ (2-shot)\end{tabular}}} & \begin{tabular}[c]{@{}l@{}}Direct Prompting\end{tabular} & 79.54 & 81.0 & 79.77 \\
 & TopK Event Partisan View & \multicolumn{1}{l}{80.45} & \multicolumn{1}{l}{81.71} & 80.78 \\
 & +Metadata & \multicolumn{1}{l}{81.62} & \multicolumn{1}{l}{82.62} & \textbf{81.93} \\ \hline
\end{tabular}%
}
\small \caption{\small Averaged results for ideology classification across $4$ issues. \textit{TopK Event Partisan View} effectively captures event-level ideological discourse.}
\label{tab:ideology-classification-stats}
\end{table}
\vspace{-2pt}


\subsection{Partisan Classification}
\label{subsec:partisanEval}
\textbf{Task Setup.} Within each \texttt{PTP} cluster, this task focuses on evaluating whether the \textit{left-leaning} viewpoints are supported by left-biased news articles and not by right-biased ones, and vice-versa. To achieve this, we first identify the news articles corresponding to the \textit{talking points} within each \texttt{PTP} cluster. Then, for a \texttt{PTP}, given a news article along with the \textit{left-} and \textit{right-leaning} viewpoints for that \texttt{PTP}, we prompt an LLM to assign the article to the viewpoint it most closely aligns with. If the model successfully categorizes left-biased news articles to \textit{left-leaning} viewpoints, then it suggests that these viewpoints are more aligned with how \textit{left-} political ideology discusses the \texttt{PTP} (similarity for the \textit{right}). As discussed in Section~\ref{subsec:ideology-classification-task-full}, we replace the terms \textit{\{left, right\}} with \textit{\{summary1, summary2\}} in this task. This substitution helps mitigate potential bias in the LLM. The prompt template used for this process is provided in Figure~\ref{fig:promptPCC} in the Appendix.

\paragraph{Comparison Methods.} We benchmark the task performance using ChatGPT against the following methods: \textbf{1. Partisan Perspectives}, represents the perspectives generated by our framework at each \texttt{PTP}. \textbf{2. Partisan Perspectives + Metadata}, adds metadata aggregated from the top 50\% of \textit{talking points} closest to \texttt{PTP}. \textbf{3. Topically Relevant Points (TRPs)}$-$We construct a baseline based on the talking points used in the clustering process prior to generating \textit{left-} and \textit{right-leaning} interpretations of a \texttt{PTP} (Section~\ref{subsubsec:clustering}). First, we label each talking point based on the ideology label of the news article it originates from. Then, for each political ideology, we consider the top-$3$ talking points closest to the \texttt{PTP} label to obtain \textit{left-} and \textit{right-} viewpoints for that \texttt{PTP} (collectively denoted as \textbf{TRPs}).
\paragraph{Results.} Table~\ref{tab:partisanClassificationTask} shows the results of partisan classification task, with issue-specific details available in Appendix~\ref{app:partisan_classification}. We observe that \textit{Partisan Perspectives} outperforms the baseline by $+11$ F1-score points, indicating that \textit{left-} and \textit{right-leaning} viewpoints tend to be strongly aligned with news articles from their respective ideologies within each \texttt{PTP} cluster. Adding metadata to \textit{Partisan Perspectives} further enhances performance on the task. However, \textbf{TRPs} themselves do not inherently encode ideology-specific information. As a result, the LLM struggles to correctly assign news articles to their respective viewpoints.

\begin{table}[h!]
\small 
\begin{center}
\begin{tabular}{|c|p{0.55cm}|c|c|}
  \hline
{\textbf{\small Approach}} & {\textbf{\small Prec.}} & {\textbf{\small Recall}} & {\textbf{\small F1}}\\
 \hline
  
 \small TRPs (baseline) & \small 73.44 & \small 73.33 & \small 73.37 \\
  \small Partisan Perspectives  & \small 85.03 & \small 84.61 & \small 84.76 \\
  \small Partisan Perspectives + Metadata & \small \textbf{85.93} & \small \textbf{86.14} & \small \textbf{85.98} \\
  \hline 
  
\end{tabular}
\small \caption{\small Averaged results for \textit{partisan classification task} across all issues. At \texttt{PTP} cluster level, \textit{partisan perspectives} align with how each political ideology interprets the \texttt{PTP}.}
\label{tab:partisanClassificationTask}
\end{center}
\vspace{-15pt}
\end{table}

\subsection{Human Evaluation}
\vspace{-1.5pt}
We conduct a human evaluation on a small sample of data collected from $3$ randomly selected events (see Table~\ref{tab:humanEvalEvents} in the appendix for event description details). Our evaluation focuses on measuring the quality of the generated \textit{partisan perspectives} using: \textbf{\textit{summary coherence}} and \textbf{\textit{mapping quality}}. 

(1) \textbf{Summary Coherence:} For each \texttt{PTP}, we evaluate whether the \textit{left-leaning} viewpoints accurately capture key ideas from the top-$K$ left-biased \textit{talking points} in that cluster (denoted as \textit{L-Coherence}). \textit{R-Coherence} is defined in a similar manner. The goal is to ensure that the generated viewpoints are not entirely random, but consistent with the set of \textit{talking points} used to construct it. (2) \textbf{Mapping Quality:} For each \texttt{PTP}, this criterion validates whether the generated \textit{left-leaning} viewpoints are expressed in left-biased \textit{news articles} within that \texttt{PTP} cluster$-$ denoted as \textit{L-MQ}. \textit{R-MQ} is defined in a similar manner for the \textit{right-leaning} viewpoints. For mapping quality, annotators must compare the generated viewpoints with excerpts from news articles to determine if the viewpoints are reflected in the article text. In total, our evaluation covered \textit{left-} and \textit{right-leaning} viewpoints from $84$ \texttt{PTP} clusters across the $3$ selected events.

We note that evaluating \textit{mapping quality} requires reading lengthy article excerpts and thus, scaling it up is challenging. To address this, we also evaluated \textit{mapping quality} using LLM and human validation (denoted as \textbf{\textit{MQ\_LLM}}). Specifically, we prompted GPT-4o with $4$ predefined questions designed to extract evidence from news articles supporting the presence of the generated viewpoints. These questions focus on the topic, key entities and their associated sentiments. Instead of manually reviewing long excerpts, human annotators compare generated viewpoints and extracted evidence to validate if the evidence aligns with the viewpoints. Using this approach, we annotated $92$ article-viewpoint pairs for the \textit{Climate Change}-related news event. Detailed discussion on the evaluation setup for all the metrics is provided in Appendix~\ref{app:subsec:human_eval}. 

\paragraph{Discussion.} Table~\ref{tab:humanEvalQuant} shows the results for both \textit{mapping quality} and \textit{summary coherence}. Our findings indicate that generated viewpoints, for both political ideologies, have high coherence scores, suggesting that they align well with \textit{talking points} from which they were created. Further, high mapping quality scores suggest that these viewpoints are accurately reflected in the news articles. However, a manual review suggests that these viewpoints can be sometimes incorrect. These typically occurs when the LLM produces poor summaries of news articles, leading to inaccuracies in generated viewpoints (as shown in Figure~\ref{fig:negExampleIncoherency}). In addition, the LLM occasionally overlooks information cited in the news articles, leading to incorrect summaries. 

Table~\ref{tab:MQ_GPT_humanEval} shows MQ\_LLM results. We report the \textit{evidence alignment score} $-$ proportion of instances where GPT-4o extracted evidence aligns with the generated viewpoints, according to human validation for each question type. We observe a high score between GPT-4o extracted evidence and viewpoints, indicating that the generated viewpoints are expressed in the news articles. However, we observe a relatively low score for the question type $-$ entities are viewed negatively. In this case, manual inspection shows that GPT-4o sometimes misses relevant evidence from the articles, exemplified in Table~\ref{tab:gpt4oFailExample} in the Appendix.

\begin{table}[!htb]
\small
\centering
\resizebox{\columnwidth}{!}{
\begin{tabular}{lcccc}
\hline
\multicolumn{1}{c}{\textbf{Issue}} & \textbf{L-Coherence(\%)} & \textbf{R-Coherence(\%)} & \textbf{L-MQ(\%)} & \textbf{R-MQ(\%)} \\ \hline
\textbf{Climate Change} & 85.71 & 100   & 75.00 & 76.92 \\
\textbf{Coronavirus}    & 100   & 90.90 & 90.90 & 70.00 \\
\textbf{Immigration}    & 93.33 & 100   & 84.62 & 94.44 \\ \hline
\end{tabular}}
\small \caption{\small Results from $84$ annotations for \textit{summary coherence} and \textit{mapping quality}.}
\vspace{-10pt}
\label{tab:humanEvalQuant}
\end{table}

\begin{table}[]
\centering
\small 
\resizebox{\columnwidth}{!}{%
\begin{tabular}{lc}
\hline
\multicolumn{1}{c}{\textbf{\begin{tabular}[c]{@{}c@{}}\small Question Type\end{tabular}}} & \multicolumn{1}{c}{\textbf{\small Evidence Alignment Score(\%)}} \\ \hline
\small Topic                                                                                                    & \small 97.82                              \\
\small Entities viewed negatively                                                                               & \small 80.43                              \\
\small Entities viewed positively                                                                               & \small 92.39                              \\
\small Angle of discussion                                                                                      & \small 98.91                              \\ \hline
\end{tabular}%
}
\small \caption{\small Shows the GPT-4o extracted evidence and viewpoints alignment score for $92$ article-viewpoint pairs (for \textit{Climate Change}-related event).}
\vspace{-5pt}
\label{tab:MQ_GPT_humanEval}
\end{table}


\vspace{-2pt}
\section{Broader Impact}
\begin{figure}[t!]
  \centering
  \includegraphics[scale=0.184]{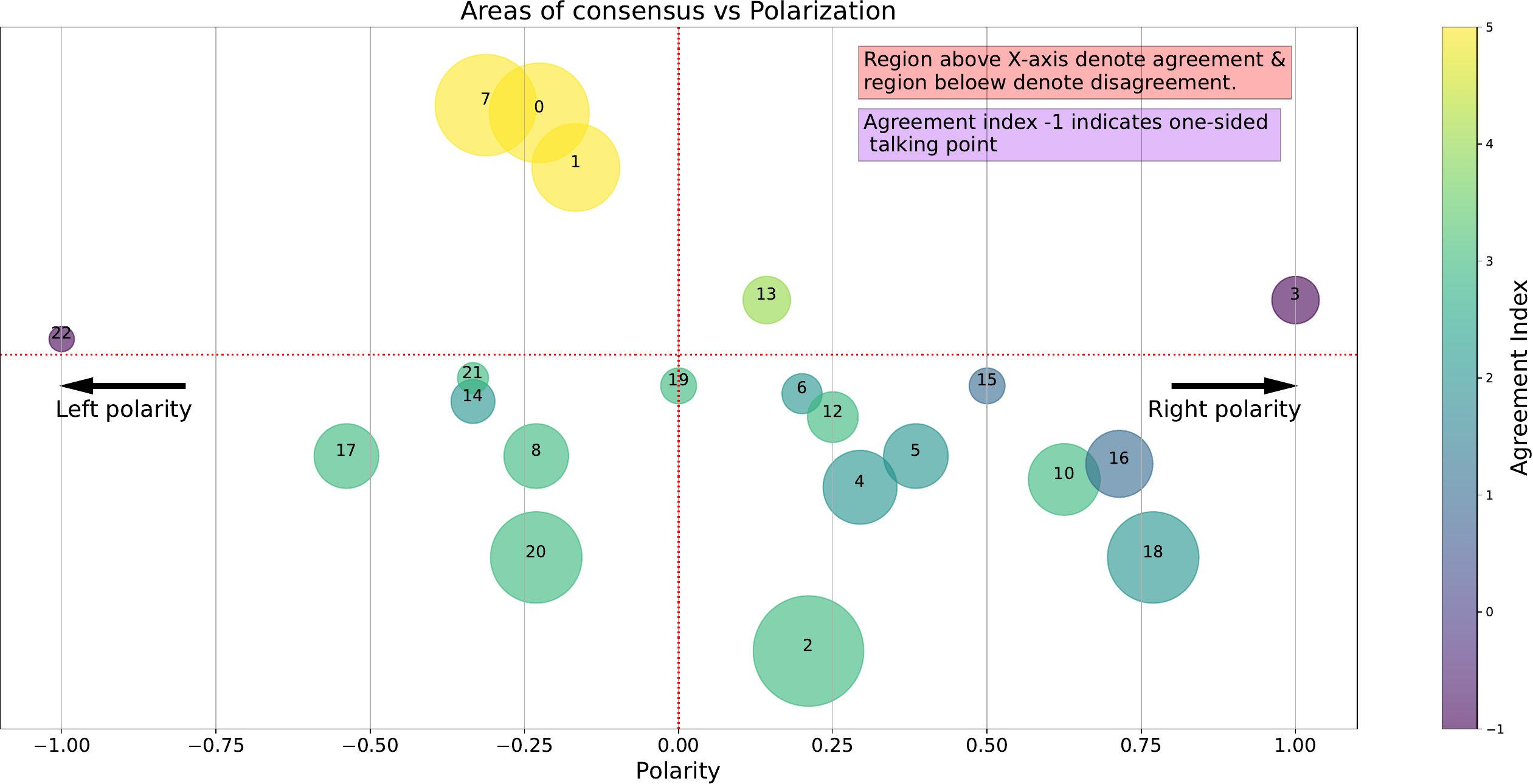}
  \small \caption{\small Discourse snapshot for a \textit{Covid}-related event. Numbers represent \texttt{PTP} IDs. Points near center denote bipartisan discussion.}
  \label{fig:partisan_narrative_visualization}
  \vspace{-15pt}
\end{figure}

\vspace{-3pt}
\paragraph{Analyzing Event-Level Discourse.} 
Our framework enables us to qualitatively visualize news events, offering insights about the \textit{areas of consensus} and \textit{polarization} by analyzing highly contested repeating themes - \texttt{PTPs}. For each \texttt{PTP}, we use GPT-4o to systematically analyze the agreement and disagreement between its \textit{left-} and \textit{right-} interpretations through a series of questions shown in the Appendix~\ref{subsec:visualPartisanNarrative}. This analysis enables us to construct an \textbf{\textit{event-discourse snapshot}}, a visualization that captures the political discourse surrounding a specific event. Figure~\ref{fig:partisan_narrative_visualization} illustrates a \textit{snapshot} for an event concerning Coronavirus - \textit{Biden's COVID-19 Relief Bill}. Our visualization maps each \texttt{PTP} as a circle along the X-axis where its position reflects its bias, with balanced view placing it in the center. Circle size indicates its frequency. Most frequent \texttt{PTPs} with high positive or negative x-axis values is evidence for "different realities", i.e., focusing on very different aspects of the topic. This visualization helps in analyzing discourse by categorizing the \texttt{PTPs} into different types $-$ \textit{agreement, disagreement, one-sided, etc.} Table~\ref{tab:narrativeViewTable} in Appendix provides an example for such categorization. For instance, talking point ID $2$ (\textit{Enhanced Vaccination Efforts}), have been discussed the most, and both political parties indicate disagreement (as it is below X-axis), signifying a strong \textit{partisan battle}. While the \textit{left} focus their discussions on \textit{equitable vaccine distribution and healthcare reform}, the \textit{right} focus mostly on \textit{criticizing Biden's vaccine distribution decisions}. This structured approach enhances the ideological discourse understanding and facilitates broader studies on media polarization and framing.

\section{Conclusion}
\vspace{-3pt}
In this paper, we introduced an LLM-based framework for analyzing ideological discourse in news events. We demonstrated the effectiveness of our approach through both automated tasks$-$ideology and partisan classification, and human validation. Additionally, we showcased the basic applications of our framework in event-discourse analysis, with plans for further investigation into its broader applicability in future work. 

\section*{Limitations and Ethics}
We identify the following limitations from our study.

\textbf{Closed-source LLM.} As a first, the perspectives generated from our framework are based on an LLM model, ChatGPT, which is closed source, and the details of its construction are unclear~\cite{spirling2023world}. We chose this model because it delivered higher quality output than other options available at the time.

\textbf{Talking point extraction.}
Our framework allows the LLM to decide the key talking points from the news article, although it is possible that it could overlook a prominent talking point. While this is a potential limitation, we believe that if a talking point is really prominent, then it will repeat in many articles, to shape the narrative. Thus, there is a high chance that the LLM would identify that talking point in other articles, even if the model failed to recognize the prominent point in the given article. 

\textbf{Labeling the talking point.}
Further, we assume that all the talking points from a left-leaning news source are actually \textit{left-biased}, and vice-versa. However, in reality, it need not be the case~\cite{kim2022measuring}. Our approach performs fairly well, even with this assumption primarily because we are only interested in identifying salient talking points from each ideology, and less frequent talking points are rejected.

\textbf{Cautious use of our framework.}
We acknowledge that our framework produces partisan viewpoints through a language model, which can occasionally generate inaccurate or misleading representations, often referred to as LLM-hallucinations. These inaccuracies may not be easily identified through automated evaluation methods. Consequently, we strongly advise users to seek expert validation before using our framework. 

While our system offers numerous practical applications, we urge users to approach its usage with caution. Although it can effectively identify ideologies, there is a potential for misuse, such as targeting individuals based on their expressed beliefs or affiliations. For these reasons, as well as many others, it is crucial for users to thoroughly evaluate both the advantages and potential risks associated with deploying our framework.

\paragraph{Ethics.}
To the best of our knowledge, we did not violate any widely held ethical precepts when producing this paper. All results are from a machine learning model and should be interpreted as such. We present all implementation and dataset details for reproducibility of our study (some parts are in the appendix). The datasets and LLMs used in this paper are publicly available and allowed for scientific research.


\bibliography{custom}

\begin{thebibliography}{51}
\expandafter\ifx\csname natexlab\endcsname\relax\def\natexlab#1{#1}\fi

\bibitem[{Aky{\"u}rek et~al.(2020)Aky{\"u}rek, Guo, Elanwar, Ishwar, Betke, and Wijaya}]{akyurek-etal-2020-multi}
Afra~Feyza Aky{\"u}rek, Lei Guo, Randa Elanwar, Prakash Ishwar, Margrit Betke, and Derry~Tanti Wijaya. 2020.
\newblock \href {https://doi.org/10.18653/v1/2020.acl-main.763} {Multi-label and multilingual news framing analysis}.
\newblock In \emph{Proceedings of the 58th Annual Meeting of the Association for Computational Linguistics}, pages 8614--8624, Online. Association for Computational Linguistics.

\bibitem[{Boydstun et~al.(2014)Boydstun, Card, Gross, Resnick, and Smith}]{Boydstun2014TrackingTD}
Amber~E. Boydstun, Dallas Card, Justin~H. Gross, Paul Resnick, and Noah~A. Smith. 2014.
\newblock \href {https://api.semanticscholar.org/CorpusID:4989256} {Tracking the development of media frames within and across policy issues}.

\bibitem[{Campello et~al.(2013)Campello, Moulavi, and Sander}]{campello2013density}
Ricardo~JGB Campello, Davoud Moulavi, and J{\"o}rg Sander. 2013.
\newblock Density-based clustering based on hierarchical density estimates.
\newblock In \emph{Pacific-Asia conference on knowledge discovery and data mining}, pages 160--172. Springer.

\bibitem[{Card et~al.(2015{\natexlab{a}})Card, Boydstun, Gross, Resnik, and Smith}]{card2015media}
Dallas Card, Amber Boydstun, Justin~H Gross, Philip Resnik, and Noah~A Smith. 2015{\natexlab{a}}.
\newblock The media frames corpus: Annotations of frames across issues.
\newblock In \emph{Proceedings of the 53rd Annual Meeting of the Association for Computational Linguistics and the 7th International Joint Conference on Natural Language Processing (Volume 2: Short Papers)}, pages 438--444.

\bibitem[{Card et~al.(2015{\natexlab{b}})Card, Boydstun, Gross, Resnik, and Smith}]{card-etal-2015-media}
Dallas Card, Amber~E. Boydstun, Justin~H. Gross, Philip Resnik, and Noah~A. Smith. 2015{\natexlab{b}}.
\newblock \href {https://doi.org/10.3115/v1/P15-2072} {The media frames corpus: Annotations of frames across issues}.
\newblock In \emph{Proceedings of the 53rd Annual Meeting of the Association for Computational Linguistics and the 7th International Joint Conference on Natural Language Processing (Volume 2: Short Papers)}, pages 438--444, Beijing, China. Association for Computational Linguistics.

\bibitem[{Chong and Druckman(2007)}]{chong2007framing}
Dennis Chong and James~N Druckman. 2007.
\newblock Framing theory.
\newblock \emph{Annu. Rev. Polit. Sci.}, 10:103--126.

\bibitem[{Dubey et~al.(2024)Dubey, Jauhri, Pandey, Kadian, Al-Dahle, Letman, Mathur, Schelten, Yang, Fan et~al.}]{dubey2024llama3herdmodels}
Abhimanyu Dubey, Abhinav Jauhri, Abhinav Pandey, Abhishek Kadian, Ahmad Al-Dahle, Aiesha Letman, Akhil Mathur, Alan Schelten, Amy Yang, Angela Fan, et~al. 2024.
\newblock \href {https://arxiv.org/abs/2407.21783} {The llama 3 herd of models}.
\newblock \emph{arXiv preprint arXiv:2407.21783}.

\bibitem[{Dubois and Blank(2018)}]{dubois2018echo}
Elizabeth Dubois and Grant Blank. 2018.
\newblock The echo chamber is overstated: the moderating effect of political interest and diverse media.
\newblock \emph{Information, Communication \& Society}, 21(5):729--745.

\bibitem[{Entman(1993)}]{Entman}
Robert~M. Entman. 1993.
\newblock \href {https://doi.org/10.1111/j.1460-2466.1993.tb01304.x} {{Framing: Toward Clarification of a Fractured Paradigm}}.
\newblock \emph{Journal of Communication}, 43(4):51--58.

\bibitem[{Fan et~al.(2019)Fan, White, Sharma, Su, Choubey, Huang, and Wang}]{fan-etal-2019-plain}
Lisa Fan, Marshall White, Eva Sharma, Ruisi Su, Prafulla~Kumar Choubey, Ruihong Huang, and Lu~Wang. 2019.
\newblock \href {https://doi.org/10.18653/v1/D19-1664} {In plain sight: Media bias through the lens of factual reporting}.
\newblock In \emph{Proceedings of the 2019 Conference on Empirical Methods in Natural Language Processing and the 9th International Joint Conference on Natural Language Processing (EMNLP-IJCNLP)}, pages 6343--6349, Hong Kong, China. Association for Computational Linguistics.

\bibitem[{Field et~al.(2018)Field, Kliger, Wintner, Pan, Jurafsky, and Tsvetkov}]{field2018framing}
Anjalie Field, Doron Kliger, Shuly Wintner, Jennifer Pan, Dan Jurafsky, and Yulia Tsvetkov. 2018.
\newblock Framing and agenda-setting in russian news: a computational analysis of intricate political strategies.
\newblock In \emph{2018 Conference on Empirical Methods in Natural Language Processing (EMNLP)}.

\bibitem[{Frermann et~al.(2023)Frermann, Li, Khanehzar, and Mikolajczak}]{Frermann_2023}
Lea Frermann, Jiatong Li, Shima Khanehzar, and Gosia Mikolajczak. 2023.
\newblock \href {https://doi.org/10.18653/v1/2023.acl-long.486} {Conflicts, villains, resolutions: Towards models of narrative media framing}.
\newblock In \emph{Proceedings of the 61st Annual Meeting of the Association for Computational Linguistics (Volume 1: Long Papers)}. Association for Computational Linguistics.

\bibitem[{Garimella et~al.(2018)Garimella, De~Francisci~Morales, Gionis, and Mathioudakis}]{garimella2018political}
Kiran Garimella, Gianmarco De~Francisci~Morales, Aristides Gionis, and Michael Mathioudakis. 2018.
\newblock Political discourse on social media: Echo chambers, gatekeepers, and the price of bipartisanship.
\newblock In \emph{Proceedings of the 2018 World Wide Web Conference}, pages 913--922. International World Wide Web Conferences Steering Committee.

\bibitem[{Gentzkow and Shapiro(2006)}]{gentzkow2006media}
Matthew Gentzkow and Jesse~M Shapiro. 2006.
\newblock Media bias and reputation.
\newblock \emph{Journal of political Economy}, 114(2):280--316.

\bibitem[{Gentzkow and Shapiro(2011)}]{gentzkow2011ideological}
Matthew Gentzkow and Jesse~M Shapiro. 2011.
\newblock Ideological segregation online and offline.
\newblock \emph{The Quarterly Journal of Economics}, 126(4):1799--1839.

\bibitem[{Goffman(1974)}]{goffman1974frame}
Erving Goffman. 1974.
\newblock \emph{Frame analysis: An essay on the organization of experience.}
\newblock Harvard University Press.

\bibitem[{Horne et~al.(2022)Horne, Gruppi, and Adali}]{nela:DVN/RBKVBM_2022}
Benjamin Horne, Mauricio Gruppi, and Sibel Adali. 2022.
\newblock \href {https://doi.org/10.7910/DVN/RBKVBM} {{NELA-GT-2021}}.

\bibitem[{Hoyle et~al.(2023)Hoyle, Sarkar, Goel, and Resnik}]{hoyle2023natural}
Alexander Hoyle, Rupak Sarkar, Pranav Goel, and Philip Resnik. 2023.
\newblock Natural language decompositions of implicit content enable better text representations.
\newblock In \emph{Proceedings of the 2023 Conference on Empirical Methods in Natural Language Processing}, pages 13188--13214.

\bibitem[{Hu et~al.(2021)Hu, Shen, Wallis, Allen-Zhu, Li, Wang, Wang, and Chen}]{hu2021lora}
Edward~J Hu, Yelong Shen, Phillip Wallis, Zeyuan Allen-Zhu, Yuanzhi Li, Shean Wang, Lu~Wang, and Weizhu Chen. 2021.
\newblock Lora: Low-rank adaptation of large language models.
\newblock \emph{arXiv preprint arXiv:2106.09685}.

\bibitem[{Jamieson et~al.(2007)Jamieson, Hardy, and Romer}]{jamieson2007effectiveness}
Kathleen~Hall Jamieson, Bruce~W Hardy, and Daniel Romer. 2007.
\newblock The effectiveness of the press in serving the needs of american democracy.

\bibitem[{Ji and Smith(2017)}]{ji-smith-2017-neural}
Yangfeng Ji and Noah~A. Smith. 2017.
\newblock \href {https://doi.org/10.18653/v1/P17-1092} {Neural discourse structure for text categorization}.
\newblock In \emph{Proceedings of the 55th Annual Meeting of the Association for Computational Linguistics (Volume 1: Long Papers)}, pages 996--1005, Vancouver, Canada. Association for Computational Linguistics.

\bibitem[{Khanehzar et~al.(2021)Khanehzar, Cohn, Mikolajczak, Turpin, and Frermann}]{khanehzar-etal-2021-framing}
Shima Khanehzar, Trevor Cohn, Gosia Mikolajczak, Andrew Turpin, and Lea Frermann. 2021.
\newblock \href {https://doi.org/10.18653/v1/2021.naacl-main.174} {Framing unpacked: A semi-supervised interpretable multi-view model of media frames}.
\newblock In \emph{Proceedings of the 2021 Conference of the North American Chapter of the Association for Computational Linguistics: Human Language Technologies}, pages 2154--2166, Online. Association for Computational Linguistics.

\bibitem[{Kiesel et~al.(2019)Kiesel, Mestre, Shukla, Vincent, Adineh, Corney, Stein, and Potthast}]{kiesel-etal-2019-semeval}
Johannes Kiesel, Maria Mestre, Rishabh Shukla, Emmanuel Vincent, Payam Adineh, David Corney, Benno Stein, and Martin Potthast. 2019.
\newblock \href {https://doi.org/10.18653/v1/S19-2145} {{S}em{E}val-2019 task 4: Hyperpartisan news detection}.
\newblock In \emph{Proceedings of the 13th International Workshop on Semantic Evaluation}, pages 829--839, Minneapolis, Minnesota, USA. Association for Computational Linguistics.

\bibitem[{Kim et~al.(2022)Kim, Lelkes, and McCrain}]{kim2022measuring}
Eunji Kim, Yphtach Lelkes, and Joshua McCrain. 2022.
\newblock Measuring dynamic media bias.
\newblock \emph{Proceedings of the National Academy of Sciences}, 119(32):e2202197119.

\bibitem[{Kinder(1998)}]{kinder1998communication}
Donald~R Kinder. 1998.
\newblock Communication and opinion.
\newblock \emph{Annual review of political science}, 1(1):167--197.

\bibitem[{Lawlor and Tolley(2017)}]{IJoC6273}
Andrea Lawlor and Erin Tolley. 2017.
\newblock \href {https://ijoc.org/index.php/ijoc/article/view/6273} {Deciding who's legitimate: News media framing of immigrants and refugees}.
\newblock \emph{International Journal of Communication}, 11(0).

\bibitem[{Lee et~al.(2022)Lee, Bang, Yu, Madotto, and Fung}]{lee2022neus}
Nayeon Lee, Yejin Bang, Tiezheng Yu, Andrea Madotto, and Pascale Fung. 2022.
\newblock \href {http://arxiv.org/abs/2204.04902} {Neus: Neutral multi-news summarization for mitigating framing bias}.

\bibitem[{Li and Goldwasser(2019)}]{li-goldwasser-2019-encoding}
Chang Li and Dan Goldwasser. 2019.
\newblock \href {https://doi.org/10.18653/v1/P19-1247} {Encoding social information with graph convolutional networks for{P}olitical perspective detection in news media}.
\newblock In \emph{Proceedings of the 57th Annual Meeting of the Association for Computational Linguistics}, pages 2594--2604, Florence, Italy. Association for Computational Linguistics.

\bibitem[{Liu et~al.(2019)Liu, Guo, Mays, Betke, and Wijaya}]{liu-etal-2019-detecting}
Siyi Liu, Lei Guo, Kate Mays, Margrit Betke, and Derry~Tanti Wijaya. 2019.
\newblock \href {https://doi.org/10.18653/v1/K19-1047} {Detecting frames in news headlines and its application to analyzing news framing trends surrounding {U}.{S}. gun violence}.
\newblock In \emph{Proceedings of the 23rd Conference on Computational Natural Language Learning (CoNLL)}, pages 504--514, Hong Kong, China. Association for Computational Linguistics.

\bibitem[{Liu et~al.(2023)Liu, Zhang, Wang, Song, Roth, and Yu}]{liu2023opendomain}
Siyi Liu, Hongming Zhang, Hongwei Wang, Kaiqiang Song, Dan Roth, and Dong Yu. 2023.
\newblock \href {http://arxiv.org/abs/2305.12835} {Open-domain event graph induction for mitigating framing bias}.

\bibitem[{Liu et~al.(2022)Liu, Zhang, Wegsman, Beauchamp, and Wang}]{liu2022politics}
Y~Liu, X~Zhang, D~Wegsman, N~Beauchamp, and L~Wang. 2022.
\newblock Politics: Pretraining with same-story article comparison for ideology prediction and stance detection.
\newblock \emph{Findings of the Association for Computational Linguistics: NAACL 2022}.

\bibitem[{Luo et~al.(2020)Luo, Card, and Jurafsky}]{luo2020detecting}
Yiwei Luo, Dallas Card, and Dan Jurafsky. 2020.
\newblock Detecting stance in media on global warming.
\newblock In \emph{Findings of the Association for Computational Linguistics: EMNLP 2020}, pages 3296--3315.

\bibitem[{McCombs and Shaw(1972)}]{mccombs1972agenda}
Maxwell~E McCombs and Donald~L Shaw. 1972.
\newblock The agenda-setting function of mass media.
\newblock \emph{Public opinion quarterly}, 36(2):176--187.

\bibitem[{Mendelsohn et~al.(2021)Mendelsohn, Budak, and Jurgens}]{mendelsohn2021modeling}
Julia Mendelsohn, Ceren Budak, and David Jurgens. 2021.
\newblock Modeling framing in immigration discourse on social media.
\newblock In \emph{Proceedings of the 2021 Conference of the North American Chapter of the Association for Computational Linguistics: Human Language Technologies}, pages 2219--2263.

\bibitem[{Moulavi et~al.(2014)Moulavi, Jaskowiak, Campello, Zimek, and Sander}]{moulavi2014density}
Davoud Moulavi, Pablo~A Jaskowiak, Ricardo~JGB Campello, Arthur Zimek, and J{\"o}rg Sander. 2014.
\newblock Density-based clustering validation.
\newblock In \emph{Proceedings of the 2014 SIAM international conference on data mining}, pages 839--847. SIAM.

\bibitem[{Nakshatri et~al.(2023)Nakshatri, Liu, Chen, Roth, Goldwasser, and Hopkins}]{nakshatri2023using}
Nishanth Nakshatri, Siyi Liu, Sihao Chen, Dan Roth, Dan Goldwasser, and Daniel Hopkins. 2023.
\newblock Using llm for improving key event discovery: Temporal-guided news stream clustering with event summaries.
\newblock In \emph{Findings of the Association for Computational Linguistics: EMNLP 2023}, pages 4162--4173.

\bibitem[{Ni et~al.(2021)Ni, Qu, Lu, Dai, {\'A}brego, Ma, Zhao, Luan, Hall, Chang et~al.}]{ni2021large}
Jianmo Ni, Chen Qu, Jing Lu, Zhuyun Dai, Gustavo~Hern{\'a}ndez {\'A}brego, Ji~Ma, Vincent~Y Zhao, Yi~Luan, Keith~B Hall, Ming-Wei Chang, et~al. 2021.
\newblock Large dual encoders are generalizable retrievers.
\newblock \emph{arXiv preprint arXiv:2112.07899}.

\bibitem[{{OpenAI}(2022)}]{openai2022}
{OpenAI}. 2022.
\newblock \href {https://openai.com} {{GPT-3.5 (ChatGPT)}}.
\newblock Computer software.

\bibitem[{Park et~al.(2021)Park, Pan, and Joo}]{park2021blames}
Kunwoo Park, Zhufeng Pan, and Jungseock Joo. 2021.
\newblock Who blames or endorses whom? entity-to-entity directed sentiment extraction in news text.
\newblock In \emph{Findings of the Association for Computational Linguistics: ACL-IJCNLP 2021}, pages 4091--4102.

\bibitem[{Quattrociocchi et~al.(2016)Quattrociocchi, Scala, and Sunstein}]{quattrociocchi2016echo}
Walter Quattrociocchi, Antonio Scala, and Cass~R Sunstein. 2016.
\newblock Echo chambers on facebook.
\newblock \emph{Available at SSRN 2795110}.

\bibitem[{Rafailov et~al.(2024)Rafailov, Sharma, Mitchell, Ermon, Manning, and Finn}]{rafailov2024directpreferenceoptimizationlanguage}
Rafael Rafailov, Archit Sharma, Eric Mitchell, Stefano Ermon, Christopher~D. Manning, and Chelsea Finn. 2024.
\newblock \href {http://arxiv.org/abs/2305.18290} {Direct preference optimization: Your language model is secretly a reward model}.

\bibitem[{Rashkin et~al.(2016)Rashkin, Singh, and Choi}]{rashkin2016connotation}
Hannah Rashkin, Sameer Singh, and Yejin Choi. 2016.
\newblock Connotation frames: A data-driven investigation.
\newblock In \emph{Proceedings of the 54th Annual Meeting of the Association for Computational Linguistics (Volume 1: Long Papers)}, pages 311--321.

\bibitem[{Roy et~al.(2021)Roy, Pacheco, and Goldwasser}]{roy2021identifying}
Shamik Roy, Mar{\'\i}a~Leonor Pacheco, and Dan Goldwasser. 2021.
\newblock Identifying morality frames in political tweets using relational learning.
\newblock In \emph{Proceedings of the 2021 Conference on Empirical Methods in Natural Language Processing}, pages 9939--9958.

\bibitem[{Santurkar et~al.(2023)Santurkar, Durmus, Ladhak, Lee, Liang, and Hashimoto}]{santurkar2023whose}
Shibani Santurkar, Esin Durmus, Faisal Ladhak, Cinoo Lee, Percy Liang, and Tatsunori Hashimoto. 2023.
\newblock Whose opinions do language models reflect?
\newblock In \emph{International Conference on Machine Learning}, pages 29971--30004. PMLR.

\bibitem[{Scheufele and Tewksbury(2007)}]{scheufele2007framing}
Dietram~A Scheufele and David Tewksbury. 2007.
\newblock Framing, agenda setting, and priming: The evolution of three media effects models.
\newblock \emph{Journal of communication}, 57(1):9--20.

\bibitem[{Spinde et~al.(2021)Spinde, Rudnitckaia, Mitrović, Hamborg, Granitzer, Gipp, and Donnay}]{SPINDE2021102505}
Timo Spinde, Lada Rudnitckaia, Jelena Mitrović, Felix Hamborg, Michael Granitzer, Bela Gipp, and Karsten Donnay. 2021.
\newblock \href {https://doi.org/https://doi.org/10.1016/j.ipm.2021.102505} {Automated identification of bias inducing words in news articles using linguistic and context-oriented features}.
\newblock \emph{Information Processing \& Management}, 58(3):102505.

\bibitem[{Spirling(2023)}]{spirling2023world}
Arthur Spirling. 2023.
\newblock Why open-source generative ai models are an ethical way forward for science.
\newblock \emph{Nature}, 616:413.

\bibitem[{Stammbach et~al.(2022)Stammbach, Antoniak, and Ash}]{stammbach-etal-2022-heroes}
Dominik Stammbach, Maria Antoniak, and Elliott Ash. 2022.
\newblock \href {https://doi.org/10.18653/v1/2022.wnu-1.6} {Heroes, villains, and victims, and {GPT}-3: Automated extraction of character roles without training data}.
\newblock In \emph{Proceedings of the 4th Workshop of Narrative Understanding (WNU2022)}, pages 47--56, Seattle, United States. Association for Computational Linguistics.

\bibitem[{Van~Dijk(1996)}]{van1996ideological}
Teun~A Van~Dijk. 1996.
\newblock Ideological discourse analysis.
\newblock \emph{MOARA--Revista Eletr{\^o}nica do Programa de P{\'o}s-Gradua{\c{c}}{\~a}o em Letras ISSN: 0104-0944}, (06):13--45.

\bibitem[{Van~Dijk et~al.(1997)}]{van1997political}
Teun~A Van~Dijk et~al. 1997.
\newblock What is political discourse analysis.
\newblock \emph{Belgian journal of linguistics}, 11(1):11--52.

\bibitem[{Ziems et~al.(2024)Ziems, Held, Shaikh, Chen, Zhang, and Yang}]{ziems2024can}
Caleb Ziems, William Held, Omar Shaikh, Jiaao Chen, Zhehao Zhang, and Diyi Yang. 2024.
\newblock Can large language models transform computational social science?
\newblock \emph{Computational Linguistics}, pages 1--55.

\end{thebibliography}
\newpage
\appendix

\section{Extended Results}
\subsection{Partisan Classification}
\label{app:partisan_classification}
Table~\ref{tab:partisanClassificationTask_detailed} shows the partisan classification results for each issue and the total aggregated performance.

\begin{table*}[ht!]
{
\begin{tabular}{llccc}
\hline
\multicolumn{1}{c}{\textbf{Issue}} &
  \multicolumn{1}{c}{\textbf{Approach}} &
  \textbf{Avg. Precision} &
  \textbf{Avg. Recall} &
  \textbf{Avg. F1-score} \\ \hline
\multicolumn{1}{l|}{\multirow{3}{*}{\begin{tabular}[c]{@{}l@{}}Climate \\ Change\end{tabular}}} &
  \multicolumn{1}{l|}{Topically Relevant Points} &
  84.11 &
  84.23 &
  84.17 \\
\multicolumn{1}{l|}{}                             & \multicolumn{1}{l|}{Partisan View}             & 91.73 & 89.46 & 90.29          \\
\multicolumn{1}{l|}{}                             & \multicolumn{1}{l|}{Partisan View + Metadata}  & 92.43 & 90.86 & \textbf{91.49} \\ \hline
\multicolumn{1}{l|}{\multirow{3}{*}{\begin{tabular}[c]{@{}l@{}}Capitol \\ Insurrection\end{tabular}}} &
  \multicolumn{1}{l|}{Topically Relevant Points} &
  69.50 &
  71.62 &
  69.18 \\
\multicolumn{1}{l|}{}                             & \multicolumn{1}{l|}{Partisan View}             & 79.33 & 80.93 & \textbf{79.93} \\
\multicolumn{1}{l|}{}                             & \multicolumn{1}{l|}{Partisan View + Metadata}  & 81.04 & 78.08 & 79.12          \\ \hline
\multicolumn{1}{l|}{\multirow{3}{*}{Immigration}} & \multicolumn{1}{l|}{Topically Relevant Points} & 69.14 & 74.64 & 69.92          \\
\multicolumn{1}{l|}{}                             & \multicolumn{1}{l|}{Partisan View}             & 85.38 & 86.36 & 85.85          \\
\multicolumn{1}{l|}{}                             & \multicolumn{1}{l|}{Partisan View + Metadata}  & 88.27 & 86.17 & \textbf{87.15} \\ \hline
\multicolumn{1}{l|}{\multirow{3}{*}{Coronavirus}} & \multicolumn{1}{l|}{Topically Relevant Points} & 73.11 & 72.60 & 72.77          \\
\multicolumn{1}{l|}{}                             & \multicolumn{1}{l|}{Partisan View}             & 83.34 & 81.76 & 82.21          \\
\multicolumn{1}{l|}{}                             & \multicolumn{1}{l|}{Partisan View + Metadata}  & 83.78 & 84.20 & \textbf{83.92} \\ \hline
\multicolumn{1}{l|}{\multirow{3}{*}{\textbf{\begin{tabular}[c]{@{}l@{}}Overall\\ Performance\end{tabular}}}} &
  \multicolumn{1}{l|}{Topically Relevant Points} &
  73.44 &
  73.33 &
  73.37 \\
\multicolumn{1}{l|}{}                             & \multicolumn{1}{l|}{Partisan View}             & 85.03 & 84.61 & 84.76          \\
\multicolumn{1}{l|}{}                             & \multicolumn{1}{l|}{Partisan View + Metadata}  & 85.93 & 86.14 & \textbf{85.98} \\ \hline
\end{tabular}}
\small \caption{Averaged results for \textit{partisan classification task} shows the efficacy of partisan perspectives in capturing ideology-specific information at the cluster-level granularity.}
\label{tab:partisanClassificationTask_detailed}
\end{table*}

\subsection{Ideology Classification Task}
\label{app:ideology-classification-task}
\paragraph{Narrative-LLAMA Quality Test.} We assess whether the perspectives generated by Narrative-LLAMA are actually expressed in the input news articles. To achieve this, we use an economical automated evaluation approach with a \textit{LLAMA3-70B-instruct} model. Specifically, for all $481$ unseen input articles, we first generate high-level ideological viewpoints using Narrative-LLAMA. We then prompt LLAMA3-70B-instruct to check if these viewpoints are represented in the corresponding news articles. We use two metrics for success: \textbf{hard} and \textbf{soft}. For each article, the \textbf{hard} metric scores $1$ if \textit{all} generated viewpoints are reflected in the article (indicating no hallucination). The \textbf{soft} metric calculates the overall proportion of viewpoints conveyed in the input articles. Table~\ref{tab:win-rate-narr-llama} presents these findings per issue, revealing that on an average, in $94\%$ of cases, \textit{all} generated perspectives are included in the input news articles, as evaluated by the 70B-instruct model.
\begin{table}[!htb]
\centering
\begin{tabular}{|l|cc|}
\hline
\multirow{2}{*}{\textbf{\small Issue Name}} & \multicolumn{2}{c|}{\small \textbf{\% Success Rate}} \\ \cline{2-3} 
 & \multicolumn{1}{l}{\small \textbf{Soft}} & \multicolumn{1}{l|}{\small \textbf{Hard}} \\ \hline
\small Climate Change & \small 97.70 & \small 93.53 \\
\small Capitol Insurrection & \small 95.97 & \small 92.15 \\
\small Immigration & \small 98.28 & \small 95.73 \\
\small Coronavirus & \small 98.39 & \small 95.06 \\ \hline
\end{tabular}%
\small \caption{LLAMA-3-70B-instruct evaluates issue-specific mapping quality, finding that on average, $94\%$ of \textbf{\textit{all}} the generated viewpoints are found in the input news articles.}
\label{tab:win-rate-narr-llama}
\end{table}

\begin{figure*}[t!]
  \centering
  \includegraphics[width=1.0\textwidth]{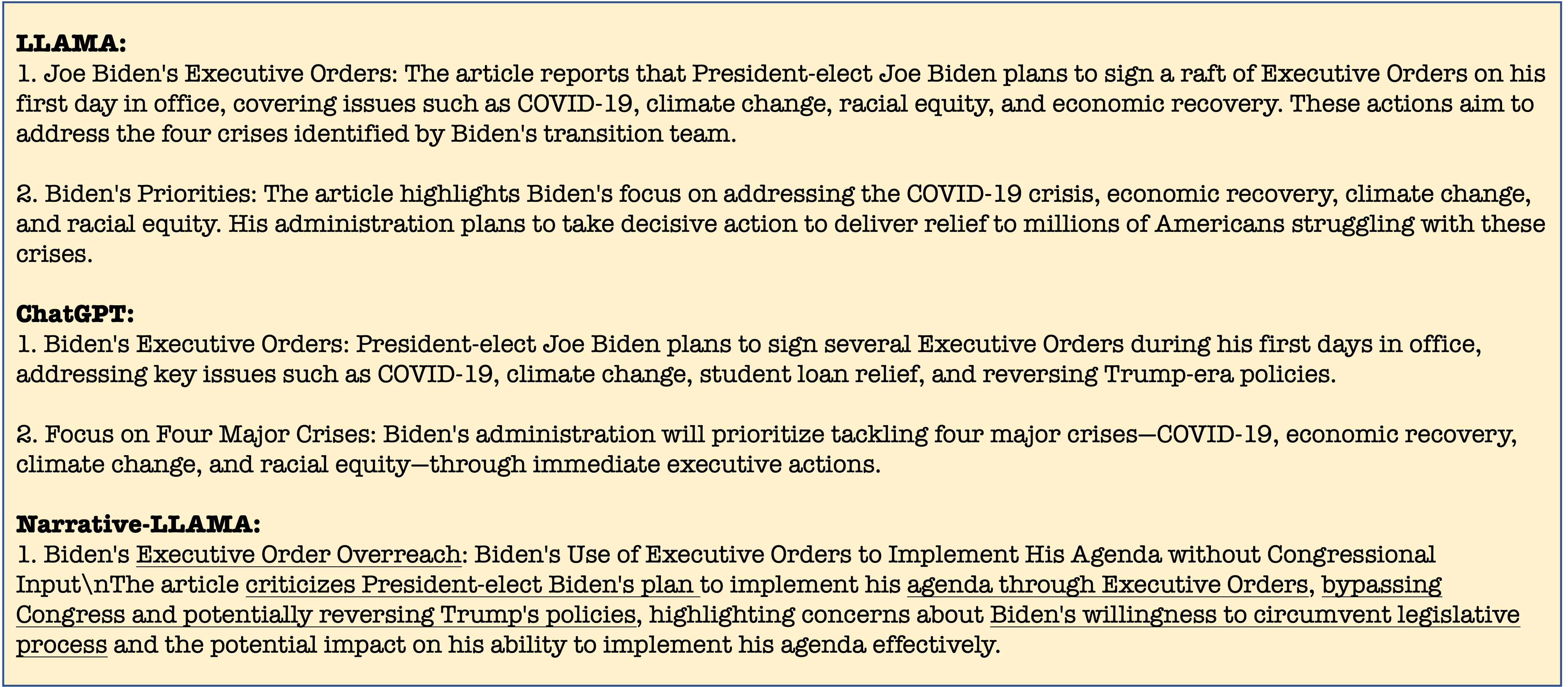}
  \small \caption{\small Shows an example of the generated perspectives for a \textit{right}-leaning news article related to \textit{Biden's executive orders}. The generated perspectives from Narrative-LLAMA capture broader ideological discourse as opposed to its counterparts.}
  \label{fig:narrative-llama-qualitative-eval}
\end{figure*}

\subsection{Evaluate clustering $-$ \texttt{PTP}}
\label{app:ptp-eval-task}
We evaluate our framework's ability to effectively cluster the talking points to form \texttt{PTPs} using two metrics $-$ \textit{coverage}, and \textit{topic diversity}. 


\paragraph{Coverage.} We define coverage as the proportion of talking points included in the clustering process. Ideally, achieving $100\%$ coverage would mean that every input talking point is fully accounted for when generating \texttt{PTPs}. Higher coverage is desirable as it ensures a more comprehensive representation of the input. Table~\ref{tab:coverage} presents the average coverage for each issue. Our results indicate that the identified \texttt{PTPs} consistently cover at least $80\%$ of the input talking points, demonstrating that they effectively represent the original set $\mathcal{T}$.


\begin{table}[!htb]
\centering
\small
\begin{tabular}{l|cc}
\hline
\multicolumn{1}{c|}{\textbf{Issue}} & \textbf{\begin{tabular}[c]{@{}c@{}}Avg. Coverage \\ per event\end{tabular}} & \textbf{Avg. \# clusters} \\ \hline
\small Climate Change       & \small 83.17 & \small 10 \\
\small Capitol Insurrection & \small 86.70 & \small 24 \\
\small Immigration          & \small 90.55 & \small 21 \\
\small Coronavirus          & \small 78.18 & \small 16 \\ \hline
\end{tabular}
\caption{\small Averaged results for coverage.}
\label{tab:coverage}
\vspace{-10pt}
\end{table}

\paragraph{Topic Diversity.}
Ideally, we want every \texttt{PTP} to be unique, as this would indicate good cluster separation. To this end, we formulate \textit{topic classification task}: Given a talking point and a set of $K'$ \texttt{PTP} labels, assign the talking point to the most topically relevant \texttt{PTP} $k^*$, where $k^* \in K'$. Note that the talking point is associated with only one of the $K'$ labels, and the rest of the labels are randomly sampled negative examples. In this case, $k^*$ helps assess how well the talking point assignments map to their respective clusters, whereas the remaining negative labels, $K' \setminus \{k^*\}$, help measure the degree of separation between the clusters.


\textbf{Setup.} We first split the data in each \texttt{PTP} cluster into 4 quartiles, where the 1st quartile refers to the top 25\% closest talking points to a \texttt{PTP} label in the embedding space, the 2nd quartile the top 50\%, etc. We randomly sample half the talking points from each quartile for this experiment. We include $3$ negative labels along with a correct label for each talking point ($|K'| = 4$). We prompt ChatGPT to assign the talking point to its most topically relevant label.


\begin{table}[!htb]
\small
\centering
\begin{tabular}{l|cccc}
\hline
\multicolumn{1}{c|}{\textbf{Issue}} & \textbf{Q1}    & \textbf{Q2}    & \multicolumn{1}{l}{\textbf{Q3}} & \multicolumn{1}{l}{\textbf{Q4}} \\ \hline
Climate Change       & 91.19 & 87.47 & 83.66 & 80.00 \\
Capitol Insurrection & 91.78 & 89.34 & 84.56 & 80.27 \\
Immigration          & 91.96 & 88.69 & 85.01 & 80.34 \\
Coronavirus          & 94.07 & 89.11 & 84.10 & 79.94 \\ \hline
\textbf{Avg. Accuracy}              & \textbf{92.74} & \textbf{88.90} & \textbf{84.37}                  & \textbf{80.12}                  \\ \hline
\end{tabular}
\caption{\small Averaged results for each quartile for the \textit{topic classification task} indicates that our prominent talking points capture diverse information.}
\label{tab:mappingQuality}
\vspace{-10pt}
\end{table}

\textbf{Discussion.} Table~\ref{tab:mappingQuality} shows the performance of the \textbf{\textit{topic classification task}}. We see that all quartiles have a decent performance, and the documents closer to the \texttt{PTP} show strongest topical relevance to the \texttt{PTP}. A strong performance from $Q4$ suggests \texttt{PTPs} capture diverse topics, and each \texttt{PTP} captures a unique aspect when compared to others.

Further, on comparing both coverage and topic diversity, we observe that \texttt{PTPs} are diverse and span at least $80\%$ of the input talking point set. This suggests that our approach forms reasoable set of \texttt{PTPs} to create \textit{partisan perspectives}.
\subsection{Human Evaluation}
\label{app:subsec:human_eval}
We perform human evaluation on a set of $3$ events. We randomly choose an event from three different issues - \{Coronavirus, Climate Change, Immigration\}. We annotate a total of $84$ \texttt{PTPs}, and describe the annotation procedure for each metric $-$ \textit{summary coherence}, and \textit{mapping quality}. Note that our annotators were graduate STEM-students who were not the authors of the paper and were under the age of 30. 

\paragraph{Summary Coherence.}
For this metric, we compare the \textit{left-} and \textit{right-} interpretations of each \texttt{PTP} label against the elements that were used to construct it. Specifically, we compare these interpretations against the top-$K$ talking points and news article summaries that were used to construct them. 

We explain the procedure for the \textit{left} political ideology, and the same process is repeated for the \textit{right} ideology as well. First, we explain the task to the annotators by providing an example. Annotators are provided with a \textit{left-leaning} viewpoint along with three to five left-biased talking points and their respective news article summaries. We ask the annotators to validate if the \textit{left-leaning} viewpoint can be derived either from the news article summaries or the talking points. If it can be derived, then the response is $1$, otherwise it is $0$. In the cases where the annotators are not sure, the response is $-1$. 

\paragraph{Mapping Quality.} For this metric, we compare \textit{left-} and \textit{right-} interpretations of each \texttt{PTP} label against the news articles to determine if they are expressed in the news articles. 

We explain the procedure for the \textit{left} political ideology, and the same process is repeated for the \textit{right} political ideology as well. In this case, we provide the annotators with a \textit{left-leaning} viewpoint and a corresponding news article that is the most relevant to the viewpoint (measured based on cosine similarity distance in the embedding space). We segment the news article into sentences of $7$, and we only provide the most relevant $7$ sentences from the news article to the annotators. First, we let the annotators know that there are at most three bullet points in the provided \textit{left-leaning} viewpoint. Then, we ask them to compare the viewpoint with the news article excerpt to validate if at least one of the points in the summary is expressed in the article. If it is, then the response is $1$, otherwise it is $0$. In cases where annotators are not sure, the response is $-1$.

\paragraph{Mapping Quality $-$ MQ\_LLM.} For this metric, human annotators validate the extracted evidence from an LLM. For the climate change event shown Table~\ref{tab:humanEvalEvents}, we examine the \texttt{PTP} clusters. From each \texttt{PTP} cluster and for each ideological-viewpoint, we randomly selected up to $5$ news articles, resulting in $92$ article-viewpoint pairs (comprising $48$ \textit{left} pairs and $44$ \textit{right} pairs). For each pair, GPT-4o was tasked with quoting relevant sentences from the articles to answer specific questions. Note that in the following questions, \textit{summary} denotes the \textit{viewpoints}.

\begin{enumerate}
    \item Is the summary discussing the same topic as the news article?
    \vspace{-10pt}
    \item In the summary and the news article, are there any entities in common that are viewed negatively from the same perspective?
    \vspace{-10pt}
    \item In the summary and the news article, are there any entities in common that are viewed positively from the same perspective?
    \vspace{-10pt}
    \item Does the news article cover the views presented in the summary from the same angle?
\end{enumerate}
Then, for each article-viewpoint pair, we present the viewpoints and the extracted evidence from the article to a human. Human is expected to validate whether the retrieved evidence aligns with the summary. Note that we proposed this experiment as an alternative evaluation scheme for mapping quality, and we limit the scale to one event considering the API cost associated with GPT-4o model.

\begin{table*}[!htb]
\small
\centering
\begin{tabular}{c|l}
\hline
\textbf{Issue} & \multicolumn{1}{c}{\textbf{News Event}} \\ \hline
\textbf{\begin{tabular}[c]{@{}c@{}}Climate \\ Change\end{tabular}} &
  \begin{tabular}[c]{@{}l@{}}\textbf{Event Title:} Biden Announces Ambitious Greenhouse Gas Emissions Cut\\ \textbf{Event Description:} This is about President Joe Biden's announcement of an ambitious cut in greenhouse gas \\ emissions as he looks to put the US back at the center of the global effort to address the climate crisis and \\ curb carbon emissions.\end{tabular} \\ \hline
\textbf{Coronavirus} &
  \begin{tabular}[c]{@{}l@{}}\textbf{Event Title:} Biden's COVID-19 Vaccination Mandate\\ \textbf{Event Description:} This is about President Joe Biden's announcement of new COVID-19 vaccination \\ requirements for federal government employees, healthcare workers, and companies with 100 or more employees, \\ and his criticism of politicians who are undermining trust in COVID vaccines.\end{tabular} \\ \hline
\textbf{Immigration} &
  \begin{tabular}[c]{@{}l@{}}\textbf{Event Title:} Biden's Refugee Cap Decision\\ \textbf{Event Description:} This is about the criticism faced by President Biden for his decision to not raise the cap \\ on refugees allowed to enter the US this year, which he had promised to do during his campaign.\end{tabular} \\ \hline
\end{tabular}
\caption{Events considered for human evaluation.}
\label{tab:humanEvalEvents}
\end{table*}

\begin{table*}[]
\centering
\begin{tabular}{p{0.55\textwidth}p{0.45\textwidth}}
\hline
\multicolumn{1}{c}{\textbf{Right-leaning viewpoints}} &
\textbf{Evidence From Article}\\
\hline
Uncertainty in global cooperation and skepticism towards US leadership. Concerns persist over the uncertainty of international support, especially from major carbon emitters like China, India, and Russia, towards America's climate initiatives. Differing views on the urgency of climate action and skepticism towards US leadership may hinder effective global collaboration on climate change.
&

Evidence: Both the viewpoints and the news article mention skepticism towards US leadership and the challenges in global cooperation. The summary states, "Concerns persist over the uncertainty of international support, especially from major carbon emitters like China, India, and Russia, towards America's climate initiatives." The news article similarly notes, "Russian President Vladimir Putin and Chinese President Xi Jinping are two notable leaders who have both confirmed their attendance at the summit, underscoring the wide range of leaders attending," indicating the importance of their participation and potential skepticism.\\
\end{tabular}
\caption{GPT-4o fails to correctly identify the relevant evidence from the news article. The negative sentiment expressed towards the countries in the viewpoints are not extracted as evidence from the news articles.}
\label{tab:gpt4oFailExample}
\end{table*}

\subsection{Visualizing Partisan Discourse in Events}
\label{subsec:visualPartisanNarrative}
To create \textbf{\textit{event-discourse snapshots}}, we would need to obtain agreement and disagreement between the \textit{left-} and \textit{right-} interpretations of each \texttt{PTP}. To obtain this, we define a scale that aids in characterizing agreement between the \textit{left-} and \textit{right-} viewpoints of a \texttt{PTP}. We begin by prompting GPT-4o to assign a binary label $-$ $0/1$, for each of the following questions. Note that in the following questions, \textit{summary} denotes the \textit{viewpoints}.
\begin{enumerate}
    \item Do both summaries have at least one common aspect of discussion?
    \vspace{-10pt}
    \item Are the summaries discussing about similar entities?
    \vspace{-10pt}
    \item Are the entities in common viewed in the same manner? For example, is the entity viewed positively or negatively in both the summaries?
    \vspace{-10pt}
    \item Do both the summaries talk about the event from the same perspective?
    \vspace{-10pt}
    \item If the summaries are viewing the event from different angles, do the summaries have atleast some agreement with each other?
\end{enumerate}

We obtained $2$ \texttt{PTPs} with a cumulative score of $1$; $5$ \texttt{PTPs} with a score of $2$; $8$ \texttt{PTPs} with a score of $3$; $4$ \texttt{PTPs} with a score of $4$; and $3$ \texttt{PTPs} with a cumulative score of $5$. We note that higher scores indicate that \textit{left-} and \textit{right-} interpretations of \texttt{PTPs} are closer to being in agreement with each other.  Lower scores imply that these are mostly disagreeing with each other. For a score of $3$, we manually inspected the outputs from the model and deduced that the two summaries shared a common aspect, discussed similar entities and had some agreements with each other. However, the entities were not viewed in the same manner due to which we assigned the ideological interpretations \texttt{PTPs} to disagree with each other. 

\begin{table*}[]
\resizebox{\textwidth}{!}{%
\begin{tabular}{ccll}
\hline
\multicolumn{1}{l}{\textbf{Categorization}} & \textbf{\texttt{PTP} ID} & \multicolumn{1}{c}{\textbf{Left-leaning Viewpoints (only titles)}}                                                                               & \multicolumn{1}{c}{\textbf{Right-leaning Viewpoints (only titles)}}                                                                                  \\ \hline
Agreement                            & 1              & Rejection of splitting COVID relief bill into separate components                & Resistance to breaking down relief package into separate bills                                                                           \\ \hline
Disagreement                         & 10             & Emphasis on Transparency and Improved Vaccine Distribution                                                                         & Questioning Biden's Vaccine Distribution Transparency                                                                                  \\ \hline
Agenda Setting                       & 16             & Biden's Travel Restrictions and Bans for Public Health                                                                             & \begin{tabular}[c]{@{}l@{}}Criticism of Biden's Executive Order on Pandemic Language\\ (for banning term - 'China Virus')\end{tabular} \\ \hline
Partisan Battle                      & 2              & \begin{tabular}[c]{@{}l@{}}Biden administration's emphasis on equitable \\ vaccine distribution and healthcare reform\end{tabular} & \begin{tabular}[c]{@{}l@{}}Criticism of Biden administration's vaccine distribution \\ decisions\end{tabular}                          \\ \hline
Right Only                           & 3              & \multicolumn{1}{c}{-}                                                                                                              & Economic Impact of \$15 Minimum Wage                                                                                                   \\ \hline
\end{tabular}%
}
\caption{Shows an example of \texttt{PTP} categorization for the \texttt{PTP} IDs shown in Figure~\ref{fig:partisan_narrative_visualization}. We show the \textit{left-leaning} and \textit{right-leaning} viewpoints for each \texttt{PTP}.}
\label{tab:narrativeViewTable}
\end{table*}

\section{Prompt Templates}
\begin{figure}[t!]
  \centering
  \includegraphics[width=1.0\columnwidth]{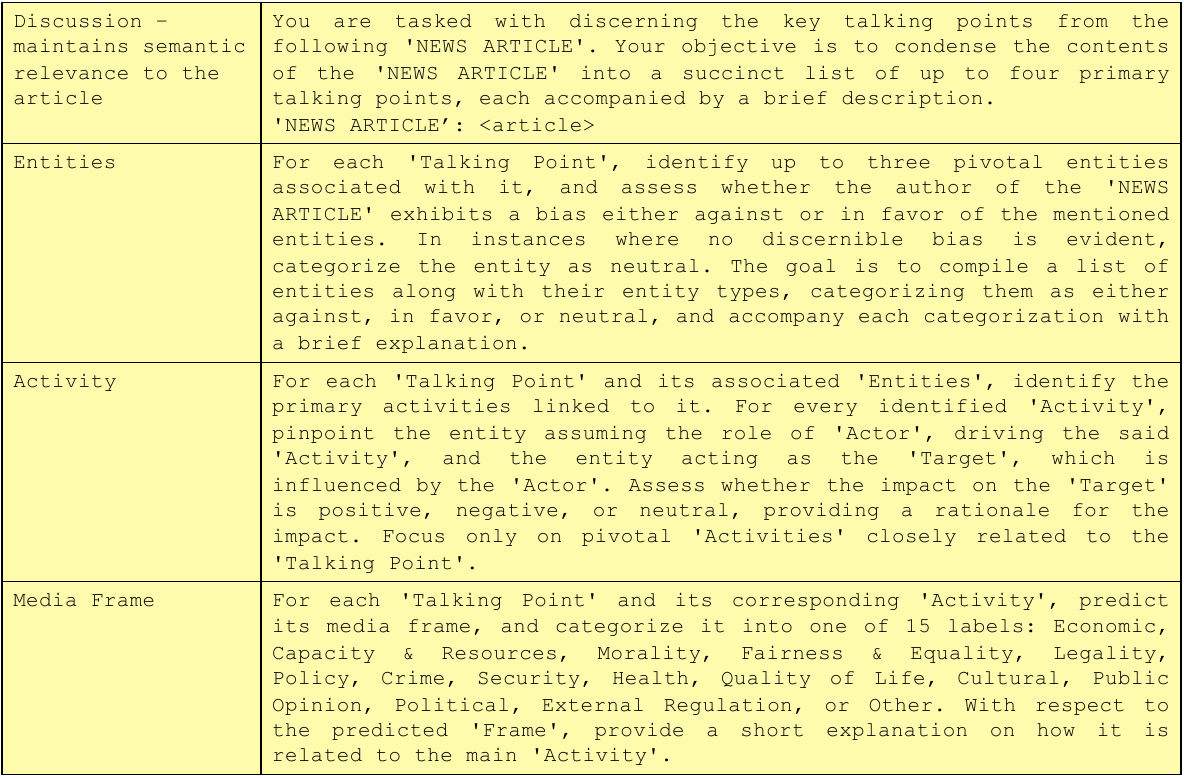}
  \small \caption{Illustrates the prompt template used to derive our relational structure, referred to as a \textit{talking point}. We begin by prompting an LLM to extract the key discussion elements $-$ repurposing the general notion of a "talking point" to identify the main topics within a news article. We then enrich this information by contextualizing it with relevant entities, their associated roles (via activity), and the surrounding media frame. In our work, this contextualized relational structure is what we define as a \textit{talking point}.}
  \label{fig:promptSchema}
\end{figure}

\begin{figure}[t!]
  \centering
  \includegraphics[width=1.0\columnwidth]{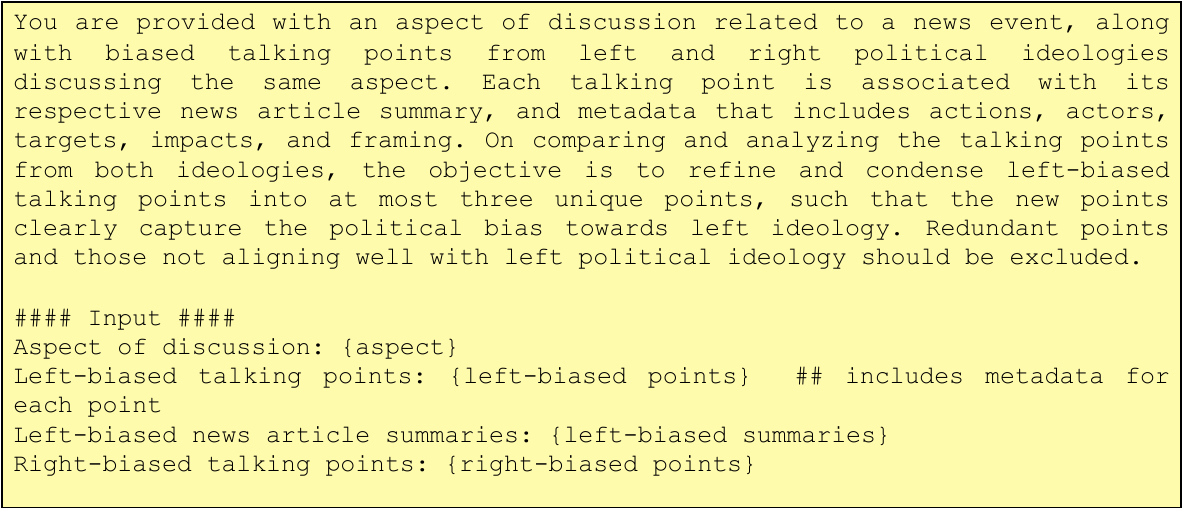}
  \small \caption{\small Shows the prompt template used to generate \textit{left-leaning} viewpoints for a \texttt{PTP}. \textit{Right-leaning} viewpoints can be obtained in a similar manner.}
  \label{fig:promptSchemaPartisanSumm}
\end{figure}

\begin{figure}[t!]
  \centering
  \includegraphics[width=1.0\columnwidth]{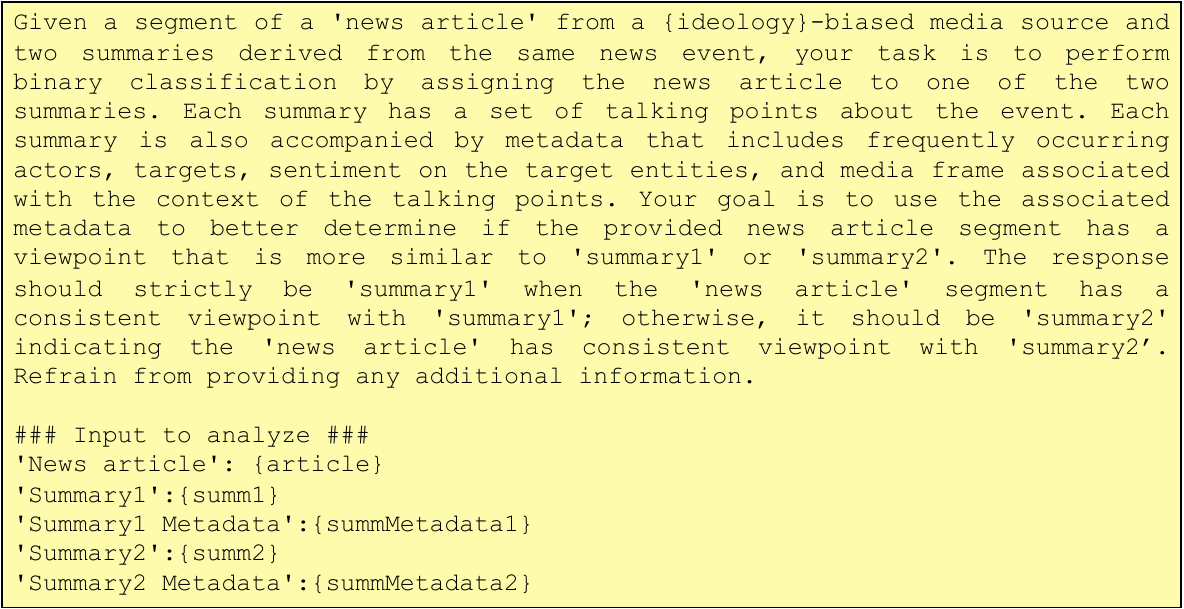}
  \small \caption{\small Shows the prompt template used for partisan classification task.}
  \label{fig:promptPCC}
\end{figure}

Figures~\ref{fig:promptSchema},~\ref{fig:promptSchemaPartisanSumm},~\ref{fig:promptPCC} shows all the prompt templates used in our work. 

\section{Experiments Related}

\subsection{Dataset extraction}
\label{app:datasetExtraction}
Here, we describe the process used for extracting the set of \textit{unseen} news articles. We note that \cite{nakshatri2023using} used NELA-2021 dataset for segmenting the news articles into a set of temporally motivated news events. In this process, \cite{nakshatri2023using} used a temporal window of $3$ in order to obtain coherent news events. 

In order to obtain \textit{unseen} news articles, yet relevant to the events under consideration, we extend this temporal window to $7$ days, and retrieve all the news articles for that time period from NELA-2021 dataset. We filter out all the news articles that part of our clustering process. Then, we consider the all the unseen articles that are closest to the event centroid in the embedding space (threshold $\geq 0.86$). Note that we obtain event centroid by averaging the embedding of all news articles relevant to the event. In this way, we extracted $481$ relevant news articles for the events under consideration, of which $234$ news articles are from right-leaning news sources, and the rest are from the left-leaning news sources.

\subsection{Narrative-LLAMA Training Details}
\label{app:narrative-llama-training-details}
Our goal is to develop a model capable of capturing higher-level ideological views within news articles, rather than focusing on minor details that are not apparent in the broader event discourse. To eliminate these minor details and enable the model to focus on the event-level ideological discourse, we train an open-sourced LLAMA3-8B model with the partisan viewpoints generated by our proposed framework. Specifically, we fine-tune the LLAMA-8B-instruct model using news article as input, and its corresponding partisan viewpoint as output. In this process, we observe that the partisan perspectives provided in training are from our framework and generated by ChatGPT, a larger LLM with superior reasoning ability compared to LLAMA3-8B. Therefore, in a way, we are distilling knowledge from a larger model (ChatGPT) to a smaller one (LLAMA3-8B), thereby instructing the smaller model to abstract away minor details and concentrate on event-level narratives. 

\paragraph{Dataset.} We select a portion of the dataset from our framework to fine-tune the model. Specifically, for each event, we consider the clusters from Section~\ref{subsubsection:partisanperspective} and extract news articles linked to the talking points in each cluster, that are the most nearest to their respective ideology-specific views of that cluster. We take the top 25\% of these articles to construct a dataset of $1100$ examples, with the news articles as inputs and their corresponding partisan viewpoints as outputs. Each input-output pair includes a news article featuring two points of view: one matching its ideological label and the other from an opposing ideology, which is crucial for training the LLM using preference optimization.

\paragraph{Training.} We follow a two-stage training process. First, we start by instruction-tuning the model using SFT, where the news article is provided as the input and the viewpoints matching the ideological label of the article serves as the output. This enables the model to adjust its output probability distribution to focus on event-level discourse. For this, we train the model using PEFT and LORA adapter parameters for $1$ epoch (with hyperparameters set to $r=64, \alpha=16, \text{\textit{learning-rate}}=1e^{-4}$). Next, we train the model using Direct Preference Optimization (DPO) for $1$ epoch, to align the model's generated outputs to prefer the viewpoints that match the ideology label of the article, while contrasting them with the viewpoints from the opposing ideology (hyperparameters: $r=64, \alpha=16, \text{\textit{learning-rate}}=5e^{-5}$)). We fine-tuned the model ($16$-bit precision) using $8$ Tesla V100 GPUs with $32$GB VRAM each, and the training process took about $3$ hours, for a batch size of $2$ per device.


\section{Clustering the talking points}
\label{app:clustering_section}
\begin{figure}[t!]
  \centering
  \includegraphics[width=1.0\columnwidth]{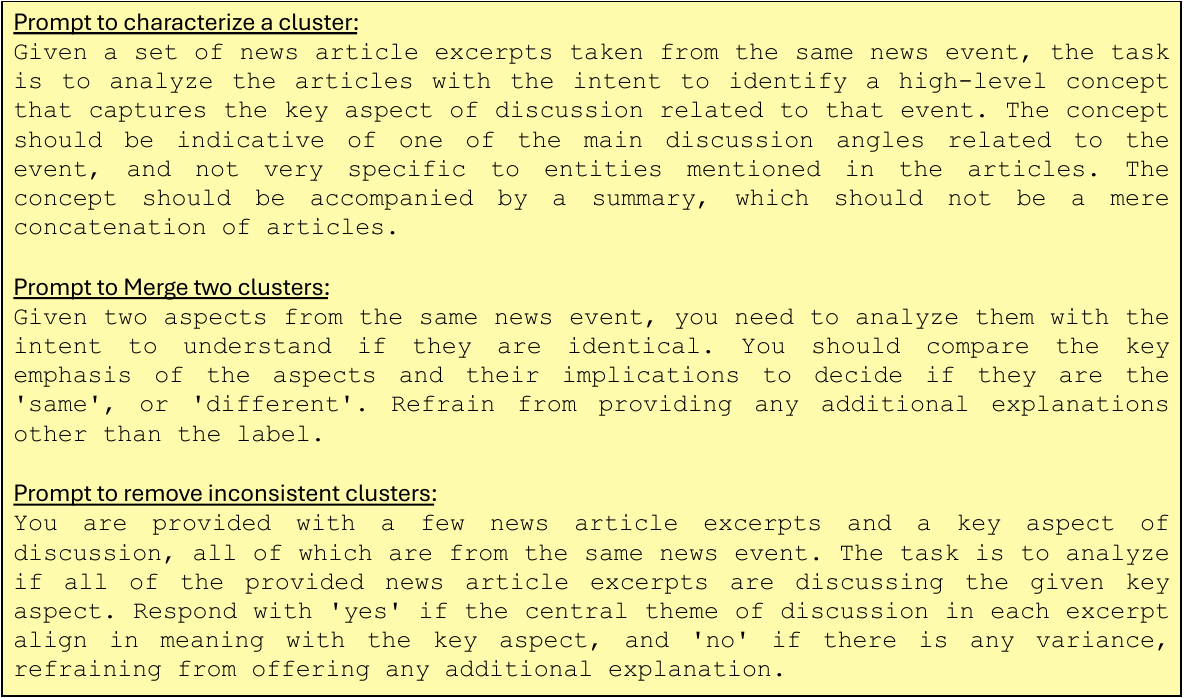}
  \small \caption{\small Shows the prompt template used to characterize a cluster, remove redundant, and incoherent clusters.}
  \label{fig:prompt-template-clustering}
\end{figure}

As described in~\ref{subsubsec:clustering}, we cluster the talking points to identify the \texttt{PTPs}. In this process, we merge redundant clusters and remove incoherent clusters. The details of this process is outlined in this section. Figure~\ref{fig:prompt-template-clustering} shows the prompt templates for the same.

\subsection{Merge Redundant Clusters}
In order to merge redundant clusters, we devise a simple greedy algorithm. We consider pairwise combinations of cluster labels, and prompt the LLM to verify if the clusters can be merged based on the prompt template shown in Figure~\ref{fig:prompt-template-clustering}. 

We start by constructing a set of pairwise cluster labels {$\mathcal{S} = \{(s_1, s_2), \cdots, (s_{n-1}, s_n)\}$}. Precisely, for every cluster, we consider all possible pairwise combinations for the top-7 closest labels to that cluster in the embedding space. For each element in $\mathcal{S}$, we prompt LLM to infer if the pair of labels are discussing about the same aspect. If the aspects, say $(s_1, s_2)$, are equivalent, then we merge these aspects, and update the set $\mathcal{S}$ by removing every element in the set that contains $s_1$ or $s_2$. In the second iteration, we construct a new set, $\mathcal{S'}$, that holds every combination of updated cluster labels, and repeat the previous step. We run the algorithm for two iterations or halt if there are no merges after the first iteration. 

\subsection{Remove Incoherent Clusters}
We note that HDBSCAN algorithm provides us with an initial set of candidate clusters. For each candidate, we use the aspect associated with the cluster label to validate if the top-3 members that are closest to the cluster label in the embedding space are discussing the same high-level concept. We prompt the LLM using the prompt shown in Figure~\ref{fig:prompt-template-clustering} to remove incoherent clusters.

\subsection{Talking Point Membership}
\label{app:membership}
After obtaining the cluster labels, which characterize the space of possible talking points. We consider each talking point from the set of all the talking points and assign the closest cluster label based on cosine similarity score. If this score is beyond a threshold value of $0.85$, we assign the talking point to that cluster label. Otherwise, the it is discarded but retained in the unclustered pool of talking points.

\subsection{Clustering-related Hyperparameters}
\label{app:cluster_hyperparam}
Note that we are interested in identifying the dense regions in the embedding space associated with talking points, as these are the potential candidate topic indicators. Due to this, we choose HDBSCAN method as our clustering algorithm, which does not require any prior number of clusters. However, we are still required to tune a few hyperparameters in order to obtain a decent performance. We use a data-driven approach to estimate the best number of topics by maximizing the DBCV score(~\citet{moulavi2014density}). We retain the default settings for \textit{cluster\_selection\_method}, and \textit{metric\_parameters}, while we change the \textit{min\_cluster\_size} and \textit{min\_samples} to get more sensible topics. This number is selected based on a grid search whose values are sensitive to the number of input talking points. Suppose |X| denote the number of talking points, then the grid parameters for HDBSCAN used in our method include {$5, 7, 9, 0.01 * |X|, 0.02 * |X|, \cdots 0.04 * |X|$}. 

For our algorithm's talking point membership module, we choose a similarity threshold of $0.76$ based on manually inspecting the prominent talking points, outputs for the cluster redundancy and removal of cluster incoherence operations for $3$ events related to the issue - \textit{Climate Change}.

\section{Examples}
\begin{figure}[t!]
  \centering
  \includegraphics[width=1.0\columnwidth]{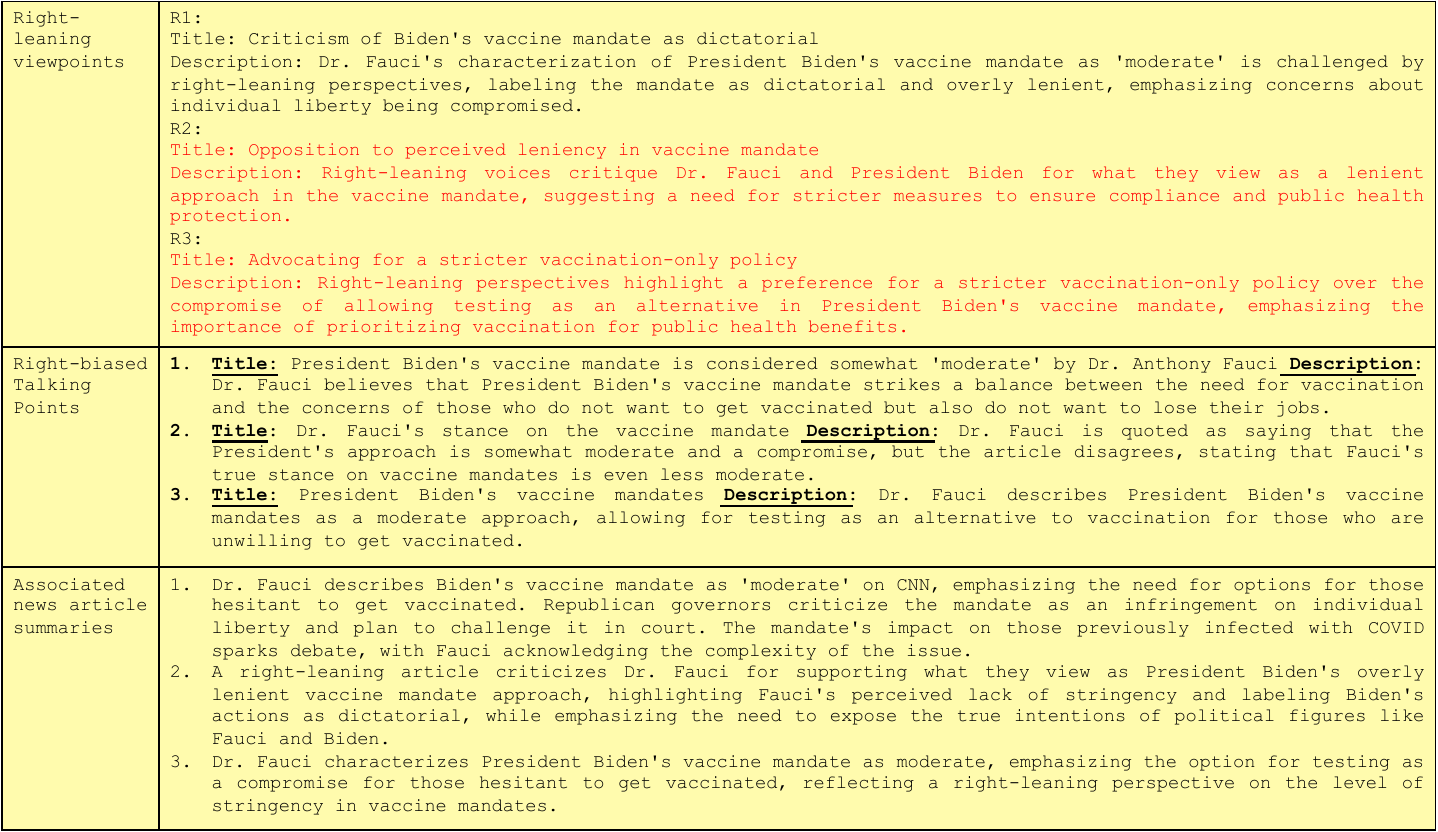}
  \small \caption{\small Shows a negative example. The generated \textit{right-leaning} viewpoints are incorrect. This is primarily attributed to inconsistent news article summaries (2 and 3), and LLM's failure to identify cited information within the news article.}
  \label{fig:negExampleIncoherency}
\end{figure}

Figure~\ref{fig:negExampleIncoherency} shows an example where the generated viewpoints are incorrect $-$ they do not align with the \textit{right} ideology views.

\section{Temporal Analysis - Case Study}
\label{appsec:temporal-case-study}
Here, we provide a simple case study to show how the data obtained from our framework could be utilized to study the \textit{left} and \textit{right} perspectives for an entire issue. To do this, we consider $7$ events related to the issue \textit{Coronavirus} at various points in time, and at every point, analyze the most frequently repeating prominent talking point from each political ideology. 

\begin{figure*}[htp]

\begin{subfigure}{\textwidth}
\includegraphics[clip,width=\textwidth,height=8cm]{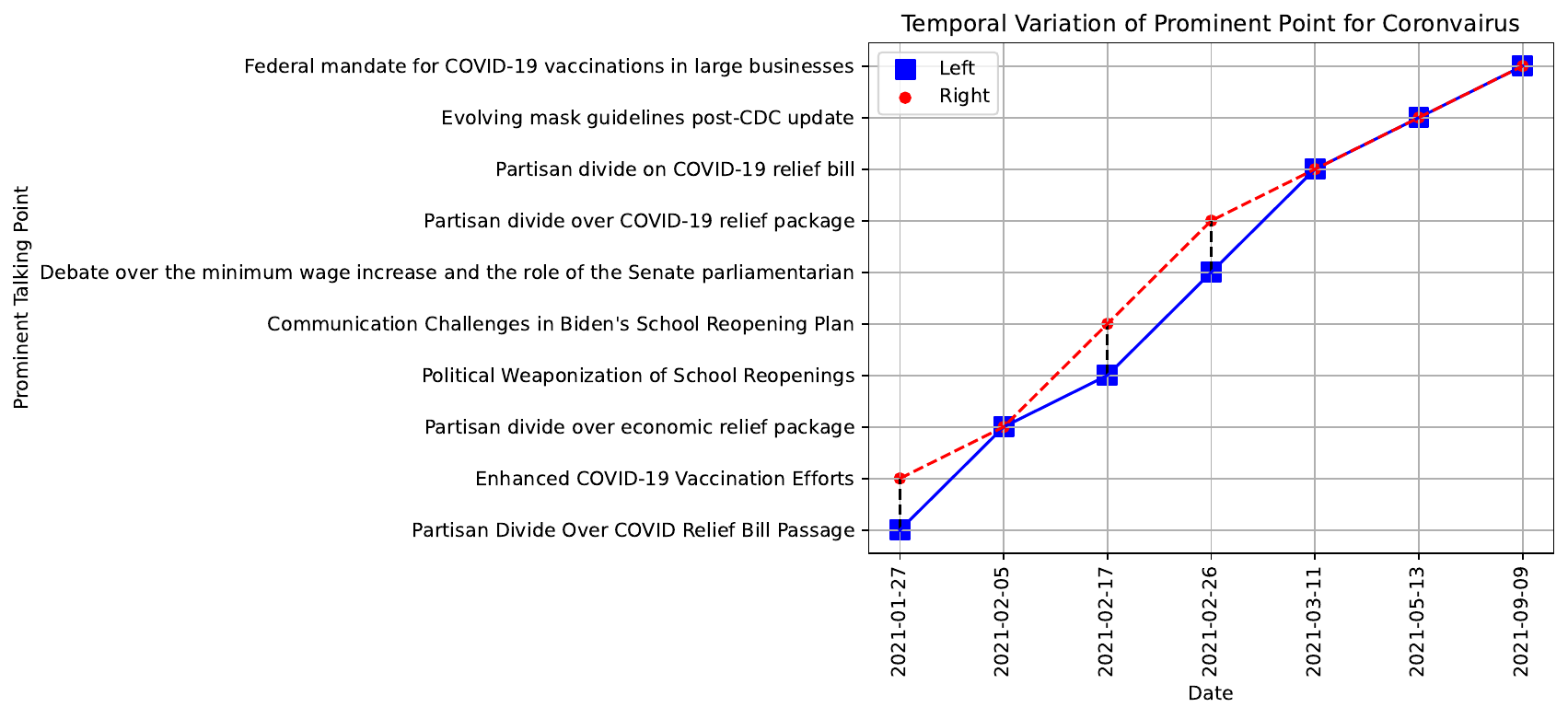}
\small \caption{\small Compares the temporal variation of most frequent prominent point for each political ideology, and across $7$ events related to the issue - \textit{Coronavirus}. X-axis indicates event-timeline, and Y-axis shows the most prominent talking point label. Frequently discussed prominent point across the two ideologies intersect in $4$ out of $7$ cases.}
\end{subfigure}

\bigskip

\begin{subfigure}{\textwidth}
\includegraphics[clip,width=\textwidth,height=10cm]{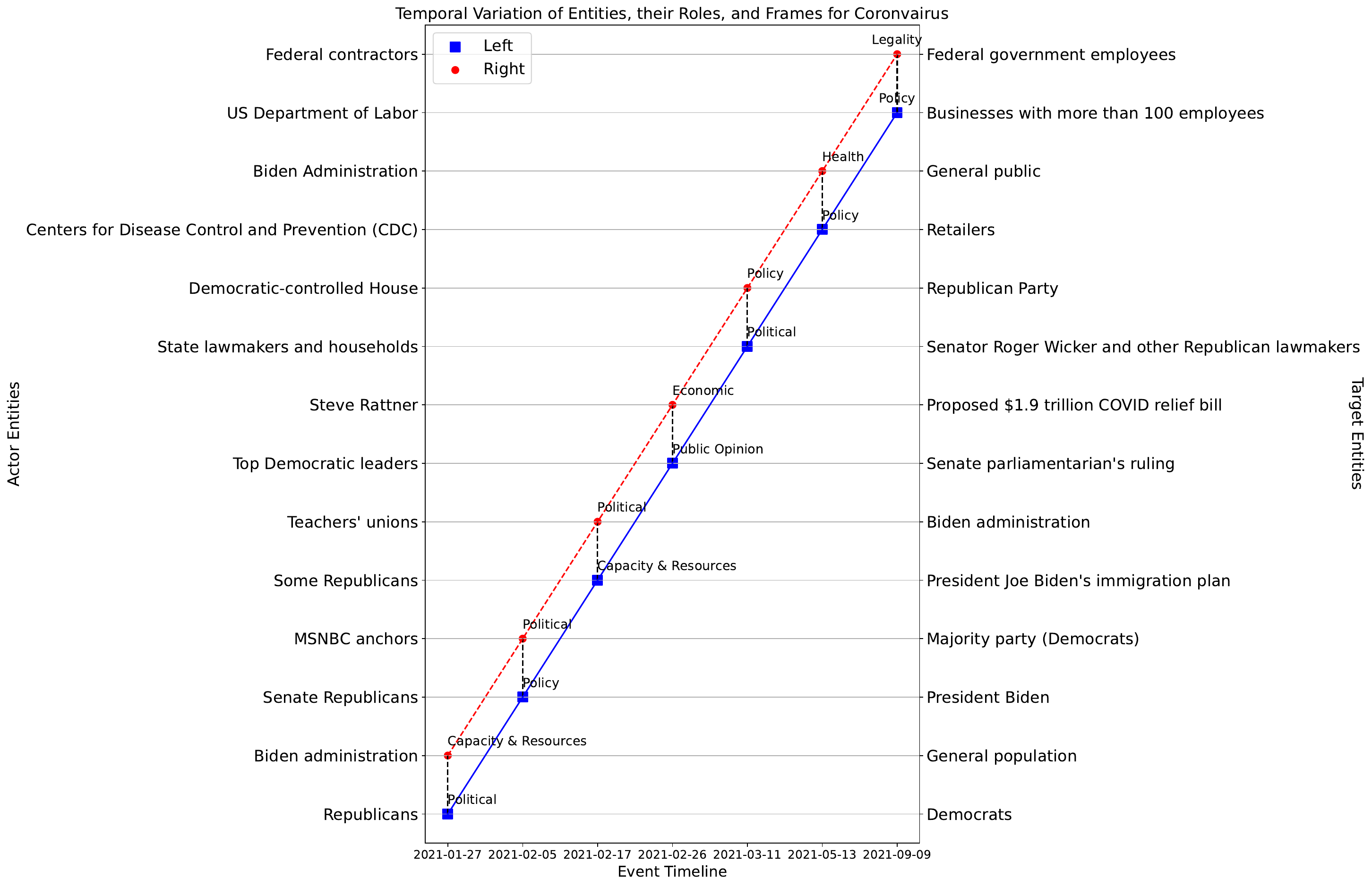}
\caption{Illustrates the temporal variation in the metadata formatted as \textit{actor-negatively impacting-target}. We show frequently repeating target entity with a negative sentiment for each ideology, and across $7$ events for the issue - \textit{Coronavirus}. For each target entity, its corresponding actor, and the associated media frame is also shown.}
\end{subfigure}

\caption{Temporal analysis of prominent points along with its respective metadata for the issue - \textit{Coronavirus}.}
\label{fig:temporalFig1}
\end{figure*}

Figure~\ref{fig:temporalFig1}(a) shows a dynamic evolution of prominent talking point of each political party for the issue - \textit{Coronavirus}. We observe that frequently discussed prominent point of each political party is different from one another in $3$ out of the $7$ events under consideration. However, both political parties predominantly discuss the same prominent point in the remaining cases. Note that Figure~\ref{fig:temporalFig1}(a) shows only the \textit{aspect} associated with each prominent point for data visualization clarity.

In the cases where both political parties discuss the same prominent point, we can further investigate the manner in which they talk about the prominent point by observing its corresponding partisan summary (\textit{left-} and \textit{right-leaning} viewpoints). For instance, let us consider the prominent point with the aspect - \textit{Evolving mask guidelines post-CDC update}, that is commonly discussed by both political parties. While both the parties criticize the ambiguity in CDC's mask guidance, the \textit{left}-leaning articles emphasizes more on pointing out the discrepancies with state and local mandates, and how it is impacting businesses. However, \textit{right}-leaning sources focus on delayed response by CDC in updating mask mandates for vaccinated individuals and raises concern about the leadership. 

We can further analyze this prominent point discussed by both parties through its associated metadata. The entity viewed as a \textit{target} by an ideology, its corresponding \textit{actor}, and the associated \textit{media frame} can help analyze the differences in the viewpoints across political parties. For the same prominent point with the aspect \textit{Evolving mask guidelines post-CDC update}, we observed that \textit{left}-leaning news sources viewed the entity \textit{Centers for Disease Control and Prevention (CDC)} to have negatively impacted the target entity \textit{Retailers}. Further investigation revealed that it was due to the criticism associated with changing mask guidelines, where \textit{CDC} removed mask mandates for the vaccinated individuals, and left-leaning sources criticized \textit{CDC} for creating ambiguity amongst the \textit{retailers} regarding the mask guidelines. We note that \textit{left}-leaning news sources commonly used \textit{Policy} as the media frame of discussion in the context of this actor-target pair. In this way, the metadata associated with the prominent point of interest can further help distinguish left and right perspectives.
To obtain an overall global view of variation in metadata for the entire issue, Figure~\ref{fig:temporalFig1} (b) shows a dynamic analysis over the actor/target entities for each prominent point across the two political parties over time.




\end{document}